%% file: main.tex
\documentclass[11pt]{article}

\usepackage[preprint]{acl}

\usepackage{times}
\usepackage{latexsym}

\usepackage[T1]{fontenc}

\usepackage[utf8]{inputenc}

\usepackage{microtype}

\usepackage{inconsolata}
\usepackage{fontawesome5}

\usepackage{graphicx}
\usepackage{amsmath,amssymb}
\usepackage{booktabs}
\usepackage{multirow}
\usepackage{stfloats}
\usepackage{wrapfig}
\usepackage{enumitem}
\usepackage{float}
\usepackage{xspace}
\usepackage[table]{xcolor}
\definecolor{rowtint}{HTML}{EAECF8}

\newcommand{\abb}{\ensuremath{\mathcal{T}}}

\newcommand{\ours}{\ensuremath{\mathcal{T}}axoSAE}
\newcommand{\edit}[1]{#1}
\definecolor{slotred}{HTML}{C00000}\definecolor{slotblue}{HTML}{00B0F0}
\newcommand{\marktext}[2]{\textcolor{#1}{#2}}

\definecolor{rowtint}{HTML}{EAECF8}
\definecolor{deltaup}{HTML}{2E8B57}
\definecolor{deltadn}{HTML}{C0392B}

\newcommand{\best}[1]{\textbf{#1}}
\newcommand{\second}[1]{\underline{#1}}

\definecolor{gainC}{HTML}{2E8B57}\definecolor{dropC}{HTML}{C0392B}

\title{\ours: Rethinking Sparse Autoencoders through Taxonomic Logic}

\author{Xinyu Guo$^{1}$ \quad Minglai Yang$^{2}$ \quad Clayton T. Morrison$^{1}$ \\
  \textbf{Eduardo Blanco$^{1}$ \quad Mihai Surdeanu$^{1}$} \\
  $^{1}$University of Arizona \quad $^{2}$Scale AI \\
  \texttt{\{xinyuguo, claytonm, eduardoblanco, msurdeanu\}@arizona.edu} \\
  \texttt{minglai.yang@scale.com}}

\allowdisplaybreaks
\begin{document}
\maketitle
\addtocontents{toc}{\protect\setcounter{tocdepth}{-1}}   
\begin{abstract}
\edit{Recent} Sparse Autoencoders (SAEs) decompose the representations of large language models (LLMs) into interpretable features, enabling precise readout and control of encoded knowledge.
\edit{However, existing SAEs treat each feature as a standalone unit, whereas real-world knowledge is often taxonomically organized.}
As a result, readout carries no guarantee of \edit{taxonomic} consistency, and control lacks the granularity to specify the scope of an edit.
We propose \ours{}, which explicitly encodes taxonomical information, built on \texttt{is-a} (hypernym–hyponym) relations, into the SAE feature space.
\ours{} mirrors a taxonomy with a stack of SAEs, one per taxonomy level, and introduces a taxonomic alignment loss that trains each concept to be active whenever one of its hyponyms is.
On two factual QA taxonomies, one over biographical relations and one over animals, \ours{} enables (1) more accurate readout through the taxonomy with concentrated activations, e.g., +31.3 points in reading out the hypernyms of abstract concepts never queried in training; \edit{(2)} more precise control, with +25.5 points in swap success, while suppressing a concept removes only its own answers (0--2\% accuracy vs.\ 91--100\% on other concepts); and, more importantly, \edit{(3)} readout of unseen concepts through their hypernyms, with up to +37.9 points in hypernym readout accuracy.
\href{https://github.com/xinnyuguo/taxo-sae}{\faGithub \textbf{ }GitHub}

\end{abstract}

\input{sections/introduction}

\input{sections/related_work}

\input{sections/task_formulation}

\input{tab_fig/fig-main-figure}

\input{sections/method}

\input{sections/experimental_setup}

\input{sections/results}

\input{sections/conclusion}

\newpage
\input{sections/limitation}

\bibliography{main}

\newpage
\appendix
\addtocontents{toc}{\protect\setcounter{tocdepth}{2}}   
\renewcommand{\contentsname}{Appendix Contents}
\makeatletter\let\taxo@lsec\l@section\renewcommand*\l@section[2]{{\renewcommand\addvspace[1]{\vskip 2pt}\taxo@lsec{#1}{#2}}}\makeatother   
\tableofcontents
\input{sections/appendix}

\end{document}

%% file: sections/introduction.tex
\section{Introduction}

Large Language Models (LLMs) encode substantial factual knowledge in their internal representations~\citep{sun2024head, cohen2023crawling,hu2025enabling} and have achieved strong performance on a wide range of knowledge-intensive tasks~\citep{yu2022generate,yifei2026researchqa}.
However, the internal representations of LLMs remain largely opaque~\citep{bereska2024mechanistic,sharkey2025open}, making it difficult to read out what knowledge they encode and to precisely control it~\citep{chang2024localization,cohen2024evaluating,belinkov2022probing}.

To mitigate this limitation, \edit{some recent} Sparse Autoencoders (SAEs) learn disentangled \edit{feature} spaces by explicitly aligning known concepts with SAE features~\citep{yang2025alignsae,harle2026measuring}. 
Similarly, contrastive activation differences~\citep{ferrando2025know,yeo2025understanding,shi2025safer} \edit{are} used with unsupervised SAEs to identify features for specific semantic information. 
These approaches enable more precise interpretation that can read out specific semantic information, and more targeted control via feature interventions.

Despite this progress, most prior work ignores dependencies between features, leading to a \textit{flat feature space where \edit{features are not logically linked}}.
This limits readout and control in realistic scenarios, where \edit{\textit{the knowledge is not flat, but often taxonomically organized.}}
For example, in biology~\citep{diehl2011hematopoietic}, knowledge about cell types follows taxonomic logic: a T cell is a type of lymphocyte, and a lymphocyte is a type of immune cell. 
In such domains, one often needs to look beyond final answers to their underlying taxonomy, e.g., checking whether an LLM identifies an unfamiliar cell type as an immune cell, and sometimes to control knowledge through this \edit{taxonomy}, e.g., suppressing knowledge of lymphocytes and all their subtypes without affecting other immune cells.

Specifically, a flat feature space limits SAEs in two ways.
First, \edit{\textit{readout carries no guarantee of taxonomic consistency}: features are isolated, so readout yields disconnected activations}, e.g., a \emph{T cell} feature may activate while the \emph{lymphocyte} feature stays silent.
Second, \textit{control lacks granularity}: a flat space has no notion of an edit's scale or scope, so one cannot specify whether an edit should be minor or substantial, or which dependent information should stay unchanged. For example, replacing \emph{T cell} with \emph{B cell} is a minor edit that keeps it a lymphocyte, whereas replacing it with \emph{neuron} is a substantial one that also changes its hypernyms, yet a flat space treats both as the same \edit{feature swap}.

To bridge this gap, we propose to {\em encode taxonomical logic 
directly into the feature space}.
\edit{We introduce \ours{} with two components:} (1) {\em Stacked SAEs}. 
\ours{} mirrors the taxonomy with a stack of SAEs, one per taxonomy level, ordered from the most specific to the most abstract concepts. The first SAE reads the LLM representation and maps it onto the most specific concepts, and each higher SAE reads the reconstruction of the SAE below and maps it onto its hypernyms. (2) {\em Taxonomic alignment loss}. To realize the \texttt{is-a} logic (i.e., hyponym-hypernym) between the stacked SAEs, we link concept activations across levels through a training loss: a concept is trained to be active whenever one of its hyponyms is. This turns the rule ``if a concept holds, its hypernym holds'' into a training objective. \edit{Further, this simplifies training considerably, as \ours\ requires explicit data points only for terminal concepts\edit{, given the taxonomy}.}

Empirically, on two taxonomies, one over biographical relations and one over animals, \ours{} yields cleaner, more readable and more controllable features than the baselines under \edit{three} evaluations. \textit{\edit{(i)} Readout:} \edit{with concentrated activations (1.0--1.7 slots per abstract concept),} readout accuracy on questions about abstract concepts never seen in training is \edit{+31.3 percentage points over} the strongest baseline. \textit{\edit{(ii)} Control:} editing the feature of a concept steers the model's answer to another concept with \edit{+25.5 percentage points in} success, and suppressing it removes only that concept's answers, with 0--2\% accuracy against 91--100\% on other concepts. \edit{\textit{\edit{(iii)} Unseen concepts:} for concepts absent from training, the upper levels still place them under the right hypernym, with up to +37.9 percentage points in accuracy.}

Our contributions are as follows:
\begin{itemize}[leftmargin=*,topsep=0.2pt,itemsep=0pt,parsep=0pt]
    \item We identify a misalignment between \edit{SAE feature spaces} and the structure of real-world knowledge: while knowledge is often \edit{taxonomically organized}, \edit{SAE features} are not aligned with it, leaving readout without \edit{taxonomic} consistency and control without granularity.
    \item We propose \ours{}, which, to the best of our knowledge, is the first to explicitly encode a target taxonomy into the SAE feature space. It mirrors a taxonomy with stacked SAEs and turns the \texttt{is-a} rule into a taxonomic alignment loss, requiring supervision only on terminal concepts. \edit{It outperforms all baselines in taxonomy readout, control.}

    \item \edit{\ours{} generalizes: it reads out unseen concepts at their hypernyms, whose \texttt{is-a} relations are absent from the SAEs' supervision; e.g., with \emph{honeybee} bound but \emph{beetle} unseen, beetle questions still activate the \emph{insect} slot, the baselines rarely read them as insect. This improves data efficiency, as not every concept needs binding, and suggests alignment with the taxonomy.}
\end{itemize}

%% file: sections/related_work.tex
\section{Related Work}
\label{sec:related}

\paragraph{Sparse Autoencoders and Concept Alignment.}
SAEs decompose LLM representations into an overcomplete dictionary of sparse features~\citep{elhage2022toy,bricken2023towards,huben2024sparse}, and have been scaled to large models~\citep{gao2025scaling} with improved activation functions~\citep{rajamanoharan2024jumping}.
Since they are trained without supervision~\citep{mudide2025efficient,bussmann2025learning,muchane2025hierarchical}, linking a feature to a concept requires post-hoc selection, e.g., by contrastive analysis of activations~\citep{yeo2025understanding,ferrando2025know}, and a single concept is often split across many features or absorbed into more specific ones~\citep{chanin2025absorption}.
Accordingly, evaluations find SAE features to be imperfect units for concept readout and control~\citep{makelov2025towards,karvonen2025saebench}.
Supervised alternatives instead bind concepts to designated units~\citep{koh2020concept}, and AlignSAE~\citep{yang2025alignsae} and G-SAE~\citep{harle2026measuring} apply this to SAEs by binding each concept to a dedicated SAE slot via a supervised loss.
\ours{} builds on this slot binding, but whereas their slots are standalone, it organizes them into levels linked by \texttt{is-a} relations.

\paragraph{Taxonomic Knowledge in LLMs.}
LLMs encode substantial factual knowledge in their internal representations~\citep{sun2024head,hu2025enabling}, much of which is organized by taxonomies and underlies tasks such as scholarly question answering~\citep{yifei2026researchqa}, cell type annotation~\citep{hou2024assessing}, and medical coding~\citep{soroush2024large}.
However, LLMs make errors on such knowledge, especially on the most specific concepts~\citep{sun2024large}.
Prior analyses show that taxonomic structure is reflected in the geometry of LLM representations~\citep{park2025geometry}, but \edit{focus} on discovering this structure rather than on reading out and controlling concepts consistently across taxonomy levels.
\ours{} makes the taxonomy explicit in the feature space, so that the concepts and hypernyms an LLM identifies can be read out, and its \edit{errors} corrected by editing them.

\paragraph{Logic Rules Encoding.}
Logic rules specify constraints that a set of facts must jointly satisfy, and integrating them into neural networks improves accuracy, data efficiency, and \edit{consistency} with background knowledge~\citep{giunchiglia2022deep,marra2024statistical}.
Prior work encodes such rules \edit{as} training objectives~\citep{hu2020harnessingdeepneuralnetworks,xu2018semantic}, \edit{constraints during} optimization~\citep{yang2023injectinglogicalconstraintsneural}, or network architecture\edit{s}~\citep{li2019augmenting}.
\edit{Following the training-objective route, our} taxonomic alignment loss encodes the \texttt{is-a} rule ``if a hyponym holds, its hypernym holds'' as a soft constraint across stacked SAEs, making taxonomic knowledge readable and editable in features.

%% file: sections/task_formulation.tex
\section{Task Formulation}
\label{sec:task}

\input{tab_fig/fig_example}
To formalize our setting, we consider taxonomy subsets represented as directed acyclic graphs $\mathcal{T}=(V,E)$ in which each node has at most one hypernym (see Fig.~\ref{fig:taxonomy-structure}). Nodes denote taxonomic elements\edit{, i.e., concepts}, and edges encode immediate hyponymy: an edge $(a,b)\in E$ indicates that $a$ is the hypernym of $b$, and $b$ is a hyponym of $a$.
We call nodes without parents roots and those without children terminal nodes; remaining nodes are intermediate. 
We index the levels of the taxonomy from the bottom: $L_1$ holds the terminal nodes and $L_K$ the roots, and $\mathcal{V}_k\subseteq V$ denotes the nodes of $L_k$.
In this work, we study factual question answering (QA)
over triples $(e,r,v)$, where a question specifies an entity
$e$ and a relation $r$, and the model returns the corresponding
value $v$.
We construct two representative taxonomies: \textbf{relation taxonomies} over $r$ and
\textbf{entity taxonomies} over $e$.
\edit{We call the element of the triple organized by the taxonomy the
\textit{taxonomic element} $t$, and the other the \textit{complementary element} $c$.}
Thus, $(t,c)=(r,e)$ for relation taxonomies and
$(t,c)=(e,r)$ for entity taxonomies.
\edit{For example, for ``How many years does a cow typically live?'' (Fig.~\ref{fig:taxonomy-structure}, gray, $K=3$), the taxonomic \edit{element} is $t=$~\emph{cow}, together with its hypernyms \emph{mammal} and \emph{vertebrate}, and the complementary \edit{element} is $c=$~\emph{typical lifespan}.}

%% file: tab_fig/fig_example.tex
\begin{figure}[t]
    \centering
    \includegraphics[width=0.85\linewidth]{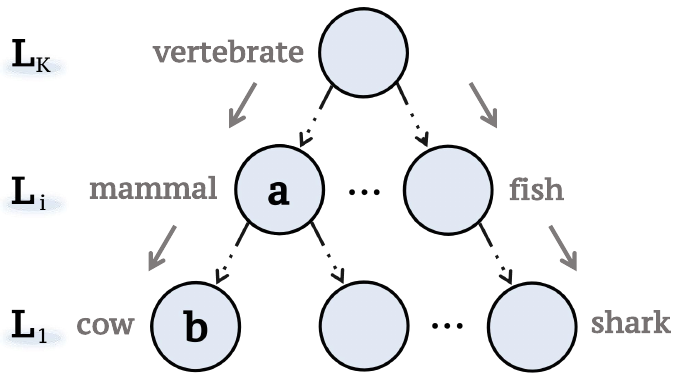}
    \vspace{-5pt}
    \caption{A taxonomy: each edge $(a,b)$ is an \texttt{is-a} link from hypernym $a$ to hyponym $b$; dashed edges mark further levels and nodes. Gray labels: a $K=3$ example from the Animal taxonomy (full taxonomy in App.~\ref{appendix:datasets}). \ours{} uses one SAE per level $L_k$ (Fig.~\ref{fig:main-figure}).}

\label{fig:taxonomy-structure}
    \vspace{-15pt}
\end{figure}

%% file: tab_fig/fig-main-figure.tex
\begin{figure*}[t]
    \centering
    \includegraphics[width=1\linewidth]{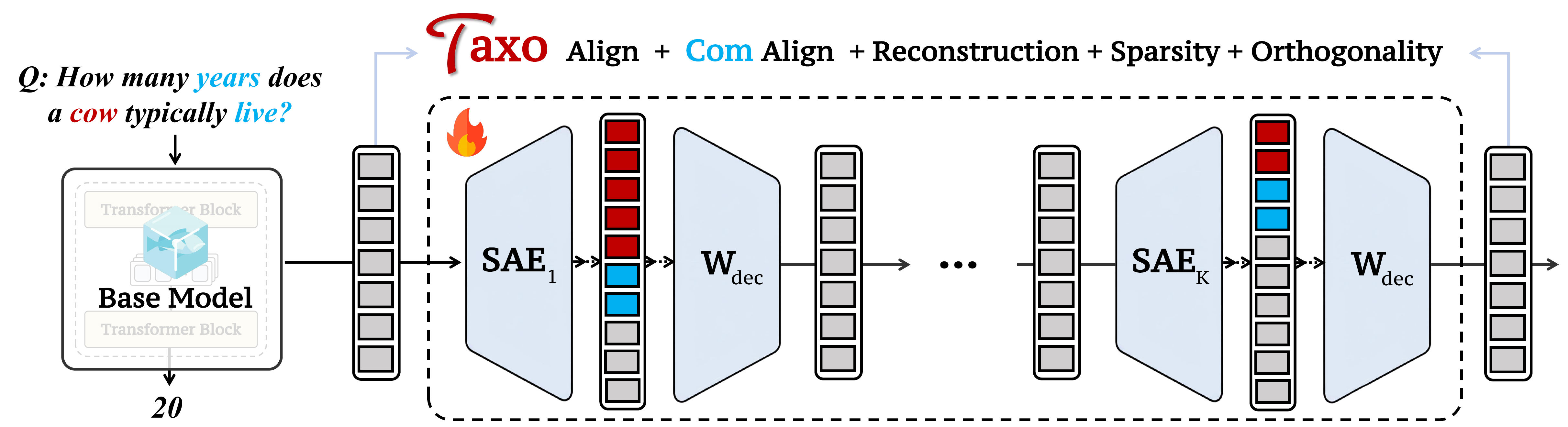}
    \vspace{-13pt}
    \caption{Overview of \ours{}. We train stacked SAEs on a frozen LLM's last-token hidden state.
    \textbf{(1) Stacked SAEs:} one SAE per taxonomy level ($L_{1,\dots,K}$ in Fig.~\ref{fig:taxonomy-structure}), each mapping the reconstruction of the level below onto its hypernyms.
    \textbf{(2) Taxonomic alignment loss:} each \marktext{slotred}{taxonomic slot} is trained to fire when the input involves its node or any of the node's hyponyms. \marktext{slotblue}{Complement slots} bind the complementary element; reconstruction, sparsity, and orthogonality losses complete the objective. Figure~\ref{fig:slot-layout} shows the full slot layout of each level on both taxonomies.
    }
    \label{fig:main-figure}
    \vspace{-13pt}
\end{figure*}

%% file: sections/method.tex
\section{Method}
\label{sec:method}

\subsection{\ours{} Architecture}

To interpret the hidden states of a frozen LLM as an explicit taxonomy, we build a stack of SAEs that encode the hidden state at a given layer of the LLM. Since the last input token summarizes the prompt ~\citep{lindsey2025biology,afzal2025knowingsayingllmrepresentations}, we follow AlignSAE~\citep{yang2025alignsae} and take the hidden state at this position before the first output token. 
Formally, given a frozen LLM $M_\theta$ with \edit{hidden dimension} $d$ and input $q=[q_1,\dots,q_n]$, the input to the SAE stack is $h^{(0)}\in\mathbb{R}^{d}$, the layer-$l$ hidden state of $M_\theta$ at the last token $q_n$ ($l$ set per dataset).

\paragraph{Stacking SAEs along the Taxonomy.}
\ours{} mirrors the taxonomy with a stack of SAEs, one SAE per taxonomy level, ordered from the terminal to the top-level nodes (Fig.~\ref{fig:main-figure}).
The $k$-th SAE, $\mathrm{SAE}_k$, models level $L_k$ of the taxonomy (\S~\ref{sec:task} and Fig.~\ref{fig:taxonomy-structure}): $\mathrm{SAE}_1$ reads the LLM hidden state and binds the terminal nodes, and each higher $\mathrm{SAE}_k$ reads the reconstruction of $\mathrm{SAE}_{k-1}$ and binds the nodes of $L_k$.
Moving up the stack thus abstracts the representation of a question from its terminal node to the node's hypernyms. Adjacent levels are connected by \edit{hyponymy relations}, which we encode as a training loss (\S~\ref{sec:method_model_training}). 
Writing $\mathrm{Enc}_k$ and $\mathrm{Dec}_k$ for the encoder and decoder of $\mathrm{SAE}_k$\edit{, $z^{(k)}$ for its feature vector, and $h^{(k)}$ for its reconstruction}, the stack is
$
h^{(0)} \xrightarrow{\mathrm{Enc}_1} z^{(1)} \xrightarrow{\mathrm{Dec}_1} h^{(1)} \xrightarrow{\mathrm{Enc}_2} \cdots \xrightarrow{\mathrm{Enc}_K} z^{(K)} \xrightarrow{\mathrm{Dec}_K} h^{(K)} .
$

\paragraph{Binding Taxonomic and Complementary Elements.}
Following AlignSAE~\citep{yang2025alignsae}, which binds each concept to a dedicated feature (\emph{slot}) through post-training, \ours{} binds taxonomic elements to slots.
A question, however, has more than one semantic dimension: it gives both the taxonomic element $t$ and the complementary element $c$ of a triple $(e,r,v)$ (\S~\ref{sec:task}).
\edit{For example, ``How many years does a cow typically live?'' asks about both the entity \emph{cow} ($t$) and the relation \emph{typical lifespan} ($c$).}
For a more flexible representation space, we bind both and separate \edit{them} by mapping the input of each $\mathrm{SAE}_k$ onto three types of slots: \emph{taxonomic slots}, \edit{one per node} at taxonomy level $L_k$; \emph{complement slots}, \edit{one per} possible \edit{complementary} element $c$; and unsupervised \emph{free slots} for the rest of the reconstruction.
Formally, $\mathrm{SAE}_k$ has a linear encoder and decoder,
\begin{align*}
z^{(k)}&=W^{(k)}_{\mathrm{enc}}h^{(k-1)}+b^{(k)}_{\mathrm{enc}},\\
h^{(k)}&=W^{(k)}_{\mathrm{dec}}z^{(k)}+b^{(k)}_{\mathrm{dec}},
\end{align*}
and its feature vector is $z^{(k)}=\big[z^{(k)}_{\mathrm{free}};z^{(k)}_{\mathrm{taxo}};z^{(k)}_{\mathrm{comp}}\big]$ with $z^{(k)}_{\mathrm{taxo}}\in\mathbb{R}^{|\mathcal{V}_k|}$, one entry per node of $L_k$, \edit{and $\dim(z^{(k)}_{\mathrm{free}})\gg\dim(z^{(k)}_{\mathrm{taxo}})+\dim(z^{(k)}_{\mathrm{comp}})$.}

\paragraph{Factual Value Prediction.}
Each $\mathrm{SAE}_k$ has a value head $\mathrm{Val}_k$ that answers the question from its own taxonomic slots and complement slots. The head ties every slot to the facts of its node, so that editing a slot changes the predicted answer. 
Since the answer depends on both elements, the head is bilinear in the two slot vectors: the taxonomic slots are mapped to a soft node embedding $e_{t}=\mathrm{softmax}(z^{(k)}_{\mathrm{taxo}})^{\top}C^{(k)}_{t}$ and the complement slots to a soft complement embedding $e_{c}=\mathrm{softmax}(z^{(k)}_{\mathrm{comp}})^{\top}C^{(k)}_{c}$, with learned tables $C^{(k)}_{t}$ and $C^{(k)}_{c}$, and the head outputs one logit for each candidate answer of $c$, $\mathrm{Val}_k(z^{(k)},c)=O_{c}\big((W_t e_{t})\odot(W_c e_{c})\big)\in\mathbb{R}^{|\mathcal{A}_c|}$, where $\mathcal{A}_c$ is the set of candidate answers of $c$ and $O_c$ is a linear output layer. 
Because a non-terminal node inherits the values of its descendants, its answer for $c$ is the set $U(u,c)$ of the values of all terminal nodes below $u$; the head is therefore multi-label, and at inference we take its top-$m_k$ answers, where $m_k$ is the largest answer-set size at $L_k$.

\subsection{\ours{ } Training}
\label{sec:method_model_training}

\paragraph{Training objectives.}
We train the encoders, decoders and value heads of all SAEs jointly. Our objective augments the standard SAE \edit{loss $\mathcal{L}_{\mathrm{SAE}}$} with (i) a taxonomic alignment loss that binds each node to its taxonomic slot ($\mathcal{L}_{\mathrm{taxo}}$), (ii) a complement alignment loss that binds each complementary element to its complement slot ($\mathcal{L}_{\mathrm{comp}}$), (iii) two decorrelation penalties that keep the taxonomic slots free of the complementary element and keep taxonomic information from leaking into the free slots ($\mathcal{L}_{\mathrm{tc}}$, $\mathcal{L}_{\mathrm{tf}}$), and (iv) a value loss that makes the slots support answer prediction ($\mathcal{L}_{\mathrm{val}}$).

Training questions are labeled {\em only} with their terminal node $u^{(1)}$ and complementary element $c$. The taxonomic slots of $\mathrm{SAE}_1$ are aligned directly to the terminal node label $u^{(1)}$. The non-terminal levels have no labels, so we derive their targets from the level below through the taxonomy.
Each hypernym collects the probabilities that $\mathrm{SAE}_k$ assigns to its hyponyms, and the hypernym with the largest total becomes the target of $\mathrm{SAE}_{k+1}$.
For example, if $\mathrm{SAE}_1$ assigns high probability to \emph{cow}, then \emph{mammal}, whose score sums the probabilities of \emph{cow}, \emph{antelope}, and \emph{bat}, has the largest total among the nodes of $L_2$ and becomes the target of $\mathrm{SAE}_2$.
{\em A hypernym slot is thus trained to be active whenever one of its hyponyms is}, which turns the rule ``if a node holds, its hypernym holds'' into a training objective.
Formally, let $p^{(k)}=\mathrm{softmax}(z^{(k)}_{\mathrm{taxo}})$ be the probabilities over the nodes of $L_k$ and $A^{(k)}\in\{0,1\}^{|\mathcal{V}_{k+1}|\times|\mathcal{V}_k|}$ the hyponymy edges between $L_k$ and $L_{k+1}$; then $A^{(k)}p^{(k)}$ sums these probabilities over the hyponyms of each node of $L_{k+1}$, and the target $\tilde u^{(k+1)}$ is its argmax (Eq.~\eqref{eq:target}), with $\tilde u^{(1)}=u^{(1)}$.
When $\mathrm{SAE}_1$ reads $u^{(1)}$ correctly, $\tilde u^{(k+1)}$ is exactly the hypernym of $u^{(1)}$ at $L_{k+1}$, which we denote $u^{(k+1)}$.
With $\mathrm{NMSE}(\hat h,h)=\|\hat h-h\|_2^2/\|h\|_2^2$, the losses of $\mathrm{SAE}_k$ \edit{(each ablated in \S~\ref{sec:ablation})} are\vspace{-5pt}
\begin{flalign}
\small
&\tilde u^{(k+1)}=\arg\max\nolimits_j\big[A^{(k)}\,p^{(k)}\big]_j, && \label{eq:target}\\
&\mathcal{L}^{(k)}_{\mathrm{SAE}}\!=\!\mathrm{NMSE}(h^{(k)}\!,h^{(k-1)})\!+\!\lambda_{\mathrm{sp}}\|z^{(k)}\|_1, &&\\
&\mathcal{L}^{(k)}_{\mathrm{taxo}}=\mathrm{CE}(z^{(k)}_{\mathrm{taxo}},\tilde u^{(k)}), &&\\
&\mathcal{L}^{(k)}_{\mathrm{comp}}=\mathrm{CE}(z^{(k)}_{\mathrm{comp}},c), &&\\
&\mathcal{L}^{(k)}_{\mathrm{tc}}=\|\mathrm{corr}(z^{(k)}_{\mathrm{taxo}},z^{(k)}_{\mathrm{comp}})\|_F^2, &&\\
&\mathcal{L}^{(k)}_{\mathrm{tf}}=\|\mathrm{cov}(z^{(k)}_{\mathrm{taxo}},z^{(k)}_{\mathrm{free}})\|_F^2, &&\\
&\mathcal{L}^{(k)}_{\mathrm{val}}=\mathrm{BCE}(\mathrm{Val}_k(z^{(k)},c),\mathbf{1}_{U(u^{(k)},c)}). &&
\end{flalign}

\vspace{-6pt}
\noindent\textbf{Pre-Training and Post-Training.}
Like AlignSAE~\citep{yang2025alignsae}, we train \ours{} in two stages. \emph{Pre-training} learns a plain dictionary with \edit{$\mathcal{L}_{\mathrm{SAE}}$} only. \emph{Post-training} starts from this checkpoint and optimizes all SAEs jointly with the full objective, which sums the losses over the levels \edit{with weights $\lambda$ (App.~\ref{appendix:ours-training})}:
\begin{equation}
\begin{split}
\mathcal{L}\!=\!\textstyle\sum_{k}\!\big[&\mathcal{L}^{(k)}_{\mathrm{SAE}}\!+\!\lambda_{\mathrm{taxo}}\mathcal{L}^{(k)}_{\mathrm{taxo}}\!+\!\lambda_{\mathrm{comp}}\mathcal{L}^{(k)}_{\mathrm{comp}}\\
&\!+\!\lambda_{\mathrm{tc}}\mathcal{L}^{(k)}_{\mathrm{tc}}\!+\!\lambda_{\mathrm{tf}}\mathcal{L}^{(k)}_{\mathrm{tf}}\!+\!\lambda_{\mathrm{val}}\mathcal{L}^{(k)}_{\mathrm{val}}\big].
\end{split}
\label{eq:total}
\end{equation}

%% file: sections/experimental_setup.tex
\section{Experimental Setup}
\label{sec:setup}

\subsection{Datasets and Baselines} 

\noindent\textbf{Datasets.} We construct two factual QA datasets: (1) \textit{Bio}, a relation taxonomy over biographical attributes of person names, whose terminal relations come from Freebase~\citep{bollacker2008freebase}, with 18 terminal, 6 intermediate, and 3 top-level nodes; and (2) \textit{Animal}, an entity taxonomy over animal names from an \texttt{IsA} subset of ConceptNet~\citep{speer2017conceptnet}, with 18 terminal, 6 intermediate, and 2 top-level nodes. Training and OOD questions differ in templates (Bio) or synonyms (Animal).
\edit{For example, ``Who is Ralph Powers's father?'' becomes ``Name Mr.\ Ralph Stephen Powers's father.'' at test time (Bio), and \emph{bat} becomes \emph{flittermouse} (Animal).}
See App.~\ref{appendix:datasets} for details and statistics.

\noindent\textbf{Baselines.}
We compare \ours{} with three baselines \edit{built on the same three-SAE stack, differing} in how they assign a slot or feature to each element:
\textit{(1) Random Unsup.\ SAE} assigns each element a fixed feature by index or at random, averaged over four assignments.
\textit{(2) DiffMean Unsup.\ SAE} uses the same SAEs and picks, for each concept, the feature in its level's SAE whose mean activation exceeds every other concept's at that level by the largest margin, via difference-in-means~\citep{gao2025scaling}.
\textit{(3) Hybrid SAE} uses the frozen terminal-level slots of \ours{} and DiffMean-selected features from unsupervised SAEs at non-terminal levels.
Unsupervised SAEs use only reconstruction and L1 losses (App.~\ref{appendix:baselines}).
\edit{A flat AlignSAE with non-terminal labels only matches our readout, swaps far worse (5.5\% vs.\ 84.7\%), and fails to keep levels consistent (0\% vs.\ 100\%; App.~\ref{appendix:flat}).}

\subsection{Implementation}

\noindent\textbf{Base Model.}
\edit{We fine-tune GPT-2 small separately on each dataset with two types of QA: (1) hypernymy QA and (2) terminal-level factual QA.
No non-terminal factual QA is used.
Fine-tuning ensures the model knows every fact and hypernymy link \edit{(settings and accuracy in App.~\ref{appendix:base-model-training})}.
We then extract its \edit{hidden states} at the final prompt token from a middle layer: layer 6 for Bio and layer 9 for Animal, chosen by terminal-level readout and swap; App.~\ref{appendix:layer-wise-performance} confirms middle layers work best.
We also repeat Bio with Qwen2.5-0.5B (App.~\ref{appendix:qwen}).}

\noindent\textbf{\ours{}.} 
We implement \ours{} \edit{as three stacked SAEs, one per level, each with taxonomic, complement, and 10k free slots, and train it on terminal-level questions only, with 100 epochs of $\mathcal{L}_{\mathrm{SAE}}$ then 500 of all losses (App.~\ref{appendix:ours-training}).}

\input{tab_fig/main_tab}
\subsection{Metrics}
\label{section:metrics}

\noindent\textbf{Taxonomy Readout (Test A).}
The SAE assigns one unit to each node: a taxonomic slot in \ours{}, a selected feature in the baselines. We ask the \edit{unedited} LLM about a node at $L_k$ and check whether the most active unit in $\mathrm{SAE}_k$ is that of the node, and in each higher $\mathrm{SAE}_{k'}$ that of its hypernym at $L_{k'}$. \textbf{Diagonal Acc} of $\mathrm{SAE}_{k'}$ averages the per-node accuracy over $L_{k'}$:
$\mathrm{Diag}=\frac{1}{|\mathcal{V}_{k'}|}\sum\nolimits_{u\in\mathcal{V}_{k'}}\mathrm{Acc}(u),$
where $\mathrm{Acc}(u)$ is the fraction of questions with node or hypernym $u$ at $L_{k'}$ that $\mathrm{SAE}_{k'}$ reads as $u$ (1 for perfect binding, $1/|\mathcal{V}_{k'}|$ at random). We read terminal-level questions at every level ($L_1{\to}L_1/L_2/L_3$), intermediate-level \edit{ones} at $L_2$ and $L_3$ ($L_2{\to}L_2/L_3$), and top-level \edit{ones} at $L_3$ ($L_3{\to}L_3$); \edit{the latter two are unseen in training.}

\noindent\textbf{Taxonomy Controllability (Tests B, C, E).}
To test whether \edit{these} units are controllable, we edit them with slot edits. We feed the LLM a terminal-level question; $\mathrm{SAE}_k$ encodes its input \edit{$h^{(k-1)}$} into the activations $z$, we edit the slot of the source node $u$ to obtain $z'$, and the LLM continues from \edit{$h'=\mathrm{Dec}_k(z')$}. A \emph{swap} to a target node $u'$ sets $z'_u=0$ and $z'_{u'}=\alpha$; a \emph{deactivation} \edit{scales} the source slot, $z'_u=(1-\beta)\,z_u$, \edit{checking} whether \edit{it removes only the node's own answers} \edit{(App.~\ref{appendix:swap-results})}.
\begin{itemize}[leftmargin=*,topsep=0pt,partopsep=0pt,itemsep=3pt,parsep=0pt]
    \item \textbf{Terminal-level swap (Test B).} We swap the terminal node in $\mathrm{SAE}_1$. \textit{Strict} requires the target's answer for the same complementary element; \textit{In-set} accepts any answer of the target node. E.g., swapping \emph{cow} for \emph{eagle} in ``What covers a cow's body?'' requires \emph{feathers} under Strict.
    \item \textbf{Intermediate-level swap (Test C).} We swap the intermediate node in $\mathrm{SAE}_2$ for one under a different top-level node. Since \edit{an intermediate node covers all its terminal hyponyms, the target's answer is the set $U(u',c)$ of their values (\S~\ref{sec:method})}, we take the top-$|U(u',c)|$ beam candidates, and \textit{target-only coverage@$|U|$} counts a swap as correct if they contain every target answer \edit{not shared by the source}. For both swaps, we also check that the SAE then reads the target's hypernyms.
    \item \textbf{Selectivity (Test E).} We deactivate the terminal node in $\mathrm{SAE}_1$ or the intermediate node in $\mathrm{SAE}_2$, with the smallest $\beta$ at which the node's own questions are answered correctly at most 5\% of the time. \textit{Own} is the exact match on the node's own questions (lower is better), and \textit{Other} on all other questions (higher is better). \edit{E.g., deactivating \emph{mammal} should break \emph{cow} questions (low Own) but not \emph{eagle} ones (high Other).}
\end{itemize}

\noindent\textbf{Activation Concentration (Test D).}
We measure how the activation of each node is spread over units. For a node $u$ of $L_k$, let $\rho_{u,j}$ be the share of the mean (positive) activation of unit $j$ of $\mathrm{SAE}_k$ on the training questions about $u$, normalized over units. \textit{EffFeat}$(u)=\exp\big(-\sum_j\rho_{u,j}\log\rho_{u,j}\big)$ is the effective number of units the node uses (1 is a single unit; lower is better), and \textit{Top1C}$(u)=\max_j\rho_{u,j}$ is the share of its strongest unit (higher is better). Both are averaged over each level's nodes.

\vspace{-2pt}

%% file: tab_fig/main_tab.tex
\begin{table*}[t]
\centering
\small
\definecolor{basegray}{HTML}{F0F0F0}
\definecolor{deltabg}{HTML}{E8F4EC}
\newcommand{\upN}[1]{{\color{gainC}\footnotesize$\uparrow$#1}}
\newcommand{\dnbN}[1]{{\color{gainC}\footnotesize$\downarrow$#1}}
\setlength{\tabcolsep}{2pt}
\newcommand{\dashsep}{\lower1.4\dp\strutbox\hbox{\color{black!22}\vbox to \dimexpr1.4\ht\strutbox+1.4\dp\strutbox\relax{\xleaders\vbox to 2.6pt{\vss\hrule width 0.45pt height 1.4pt}\vfill}}}
\renewcommand{\arraystretch}{1.4}
\resizebox{\textwidth}{!}{%
\begin{tabular}{l*{3}{w{c}{0.99cm}}!{\dashsep}*{2}{w{c}{0.99cm}}!{\dashsep}w{c}{0.99cm}*{3}{w{c}{0.98cm}}*{5}{w{c}{0.84cm}}}
\toprule
& \multicolumn{6}{c}{Diagonal Acc (\%) $\uparrow$} & \multicolumn{3}{c}{EffFeat $\downarrow$} & \multicolumn{3}{c}{Top1C (\%) $\uparrow$} & \multicolumn{2}{c}{Swap (\%) $\uparrow$} \\
\cmidrule(lr){2-7}\cmidrule(lr){8-10}\cmidrule(lr){11-13}\cmidrule(lr){14-15}
SAE & $L_1$${\to}$$L_1$ & $L_1$${\to}$$L_2$ & $L_1$${\to}$$L_3$ & $L_2$${\to}$$L_2$ & $L_2$${\to}$$L_3$ & $L_3$${\to}$$L_3$ & $L_1$ & $L_2$ & $L_3$ & $L_1$ & $L_2$ & $L_3$ & $L_1$ & $L_2$ \\
\midrule
\multicolumn{15}{l}{\emph{Bio relations} (layer 6)} \\
\rowcolor{basegray}Random Unsup.\ SAE & 4.3 & 4.2 & 3.9 & 6.7 & 3.0 & 1.4 & 360.8 & 298.0 & 283.3 & 3.5 & 3.5 & 3.1 & 2.2 & 0.0 \\
DiffMean Unsup.\ SAE & \second{77.9} & \second{82.1} & 75.4 & \second{61.4} & \second{60.7} & 63.7 & 360.8 & 298.0 & 283.3 & 3.5 & 3.5 & 3.1 & \second{81.8} & 12.2 \\
Hybrid SAE & \best{100.0} & 71.0 & \second{77.1} & 54.9 & 54.5 & \second{69.2} & \best{3.2} & \second{188.2} & \second{184.9} & \best{74.8} & \second{4.4} & \second{3.5} & \best{84.7} & \second{16.3} \\
\rowcolor{rowtint}
\textbf{\ours} & \best{100.0} & \best{100.0} & \best{100.0} & \best{69.8} & \best{92.0} & \best{87.8} & \best{3.2} & \best{1.5} & \best{1.1} & \best{74.8} & \best{91.9} & \best{99.1} & \best{84.7} & \best{41.8} \\
\rowcolor{deltabg}
$\Delta$(\abb\ $-$ Rand.) & \upN{95.7} & \upN{95.8} & \upN{96.1} & \upN{63.1} & \upN{89.0} & \upN{86.4} & \dnbN{357.6} & \dnbN{296.5} & \dnbN{282.2} & \upN{71.3} & \upN{88.4} & \upN{96.0} & \upN{82.5} & \upN{41.8} \\
\addlinespace[3pt]
\hline
\multicolumn{15}{l}{\emph{Animal entities} (layer 9)} \\
\rowcolor{basegray}Random Unsup.\ SAE & 1.9 & 3.5 & 12.4 & 2.9 & 12.5 & 12.5 & 644.8 & 415.6 & 346.9 & 1.5 & 1.3 & 1.7 & 5.3 & 9.4 \\
DiffMean Unsup.\ SAE & \second{60.7} & \second{53.1} & 50.5 & \second{22.5} & \second{64.7} & 43.2 & 644.8 & 415.6 & 346.9 & 1.5 & 1.3 & 1.7 & \second{40.5} & \second{25.5} \\
Hybrid SAE & \best{84.3} & 18.2 & \second{55.3} & 16.3 & 50.6 & \second{45.4} & \best{4.7} & \second{95.3} & \second{86.4} & \best{63.7} & \second{7.4} & \second{4.7} & \best{43.6} & 21.4 \\
\rowcolor{rowtint}
\textbf{\ours} & \best{84.3} & \best{86.9} & \best{90.3} & \best{47.9} & \best{86.6} & \best{50.6} & \best{4.7} & \best{1.7} & \best{1.0} & \best{63.7} & \best{89.0} & \best{100.0} & \best{43.6} & \best{27.6} \\
\rowcolor{deltabg} $\Delta$(\abb\ $-$ Rand.) & \upN{82.4} & \upN{83.4} & \upN{77.9} & \upN{45.0} & \upN{74.1} & \upN{38.1} & \dnbN{640.1} & \dnbN{413.9} & \dnbN{345.9} & \upN{62.2} & \upN{87.7} & \upN{98.3} & \upN{38.3} & \upN{18.2} \\
\bottomrule
\end{tabular}}
\vspace{-5pt}
\caption{
Results on Bio and Animal. $L_1$/$L_2$/$L_3$: terminal/intermediate/top levels; L$q{\to}$L$r$: level-$r$ readout on level-$q$ questions ($L_1{\to}L_r$ pools ID and OOD).
Bold/underline: best/second-best; $\Delta$: \ours{} minus Random Unsup.\ SAE.
\edit{\ours{} achieves the best readout (Diagonal Acc) \edit{and} control (Swap) on both datasets. Layer choice: \edit{App.}~\ref{appendix:layer-wise-performance}. The same conclusions hold on Qwen2.5-0.5B (App.~\ref{appendix:qwen}).}
}

\label{tab:main}
\vspace{-10pt}
\end{table*}

%% file: sections/results.tex
\section{Experimental Results}\label{sec:results}

\subsection{Taxonomy Readout}
\input{tab_fig/fig-unseen-tsne}
\input{tab_fig/fig_activation}

\noindent\textbf{\ours{} enables more accurate taxonomy readout.}
In Table~\ref{tab:main}, \ours{} matches or exceeds all baselines in Diagonal Acc on both datasets, \edit{as on Bio with Qwen2.5-0.5B (App.~\ref{appendix:qwen})}.
\edit{On terminal-level questions, it reads terminal nodes and hypernyms accurately at all levels ($L_1{\to}L_1/L_2/L_3$).}
More importantly, though trained only on terminal-level questions, it generalizes to non-terminal questions, identifying queried
elements and their hypernyms.
For $L_2{\to}L_2$ and $L_3{\to}L_3$, it beats
the strongest baseline by up to 25.4 points.
For hypernym readout ($L_2{\to}L_3$), it gains 31.3 points.
Complement slots read out almost perfectly (App.~\ref{appendix:complement}).
\edit{The advantage also holds on unseen phrasings of terminal-level questions (App.~\ref{appendix:phrasings}).}

\paragraph{\edit{\ours{} places unseen terminal concepts under their hypernyms.}}
\edit{We test 24 animals outside the trained ones (App.~\ref{appendix:unseen}).
\ours{} maps most of them into the $L_2$ region of their class (Fig.~\ref{fig:unseen-tsne}) and reads the right intermediate and top-level node for 63.9\% and 76.1\% of questions, against 26.0\% and 56.8\% for DiffMean.
Bio's 24 unseen relations show the same pattern (91.4\% vs.\ 75.9\% at $L_2$).}

\input{tab_fig/fig-frag}

\paragraph{\edit{\ours{} produces clean and concentrated activations.}}
\edit{Figure~\ref{fig:activation} visualizes how terminal-level questions activate taxonomic elements on Bio.
In the unsupervised SAE, each question activates many features, and the same features fire on questions about different elements, so no single feature marks an element.
In \ours{}, the taxonomic slots respond selectively: a question activates mainly the slot of its own element, at both $L_1$ and $L_2$.
Its binding matrices are diagonal at all three levels: questions about an element, or about any of its hyponyms, activate that element's slot and few others.
The concentration metrics in Table~\ref{tab:main}, computed on the activations without post-hoc feature selection: \ours{} reaches non-terminal EffFeat of 1.0--1.7 and Top1C of 89.0--100.0\%, \edit{i.e., each abstract concept is carried by about one taxonomic slot, which holds most of its activation, so a single slot suffices to read out or edit the concept;} whereas the activations of unsupervised SAEs are fragmented at all three levels (Fig.~\ref{fig:frag}).}

\subsection{Taxonomy Controllability}

\input{tab_fig/tab-swap-bio}

\noindent\textbf{\ours{} achieves better target-value prediction and more consistent target-hypernym readout after swapping.}
We take Bio as an example in Table~\ref{tab:swap-bio} with complete results in App.~\ref{appendix:swap-results}.
It shows that after terminal- and intermediate-level swaps, \ours{} more reliably produces the correct answer for the target taxonomic element, \edit{especially at the intermediate level (+25.5 points in target-only coverage).}
\edit{Its higher levels also read the target's hypernyms in 93.2--100.0\% of swaps, 28.9--37.3 points above the baselines.}

\input{tab_fig/fig-selectivity}
\paragraph{\ours{} selectively suppresses target answers while preserving others.}
Figure~\ref{fig:selectivity} shows the effects of deactivating
individual taxonomic slots at terminal and intermediate levels.
Across both datasets, \ours{} reduces exact-match accuracy
for the targeted element to 0.00--0.02 while retaining
0.91--1.00 accuracy for other elements.
In contrast, deactivating DiffMean-selected features also
disrupts answers for other elements, especially at the
intermediate level, where \edit{Other} accuracy is only 0.43
on Animal and 0.60 on Bio.

\input{tab_fig/fig-ablation}
\subsection{Ablation Study}
\label{sec:ablation}

\paragraph{Ablation Variants and Readout--Controllability Metrics.}
\edit{To assess each training loss, we evaluate five ablation variants on Bio, each removing one loss from \ours{}.
Readout averages all Test A diagonal accuracies ($L_1{\to}L_1/L_2/L_3$, $L_2{\to}L_2/L_3$, $L_3{\to}L_3$), weighted by question counts; controllability averages Test B strict accuracy and Test C target-only coverage on pairs from different top-level nodes, each at its best $\alpha$.
Figure~\ref{fig:ablation} plots both (upper right is better); full results for Tests A--E are provided in App.~\ref{appendix:ablation}.}

\paragraph{Removing any of the \edit{five} losses weakens \ours{}'s joint readout and steering performance.}
Figure~\ref{fig:ablation} shows that
(1) \edit{removing taxonomic alignment (\textit{w/o taxo align}) collapses both diagonal accuracy and swap success to the level of the random baseline;}
(2) removing \edit{complement alignment} or value prediction
(\edit{\textit{w/o comp align}}, \textit{w/o value}) preserves
relatively high readout accuracy but markedly reduces swap success;
and (3) removing either orthogonality loss
(\edit{\textit{w/o ortho-tc}, \textit{w/o ortho-tf}}) shows a similar
pattern, with smaller drops in swap success than other variants, while readout remains comparable
and some concentration metrics improve.
\edit{Replacing the bilinear value head with a \edit{per-slot} head mainly weakens intermediate-level swaps (App.~\ref{appendix:ablation-value-head}).}

%% file: tab_fig/fig-unseen-tsne.tex
\begin{figure}[!t]
    \centering
    \includegraphics[width=1\linewidth]{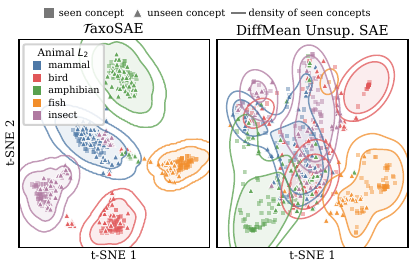}
    \vspace{-19pt}
    \caption{t-SNE of $L_2$ activations on Animal, colored by $L_2$ node: squares are trained animals, triangles unseen ones. A triangle inside its color's contour is placed under the right hypernym. \ours{} does this for most unseen animals; DiffMean's regions overlap. Mollusc is omitted due to heavy overlap in the baseline ( App.~\ref{appendix:unseen}).}
    \label{fig:unseen-tsne}
    \vspace{-10pt}
\end{figure}

%% file: tab_fig/fig_activation.tex
\begin{figure*}[t]
    \centering
    \includegraphics[width=1\linewidth]{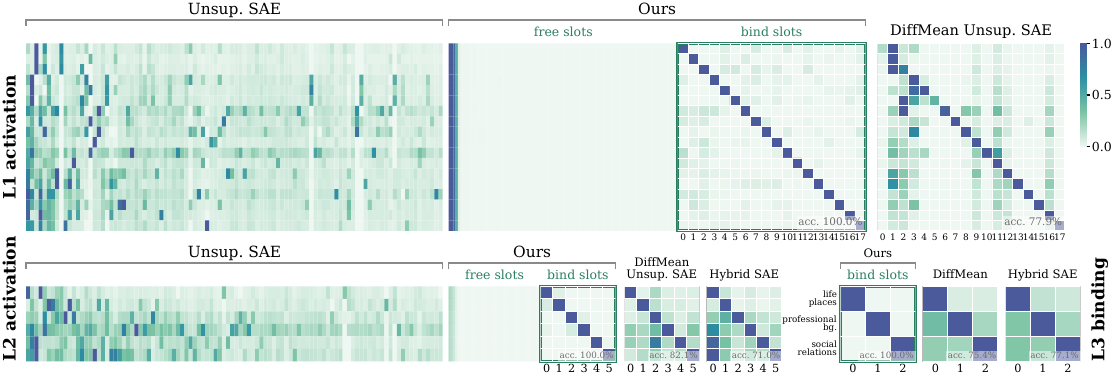}
    \vspace{-20pt}
    \caption{
Normalized activations ($L_1$/$L_2$) and binding matrices ($L_1$--$L_3$) on Bio. Rows group questions by their terminal node and its hypernyms. \edit{Wide panels show activations over all units: the unsupervised SAE spreads each concept over many features, while \ours{} concentrates it in bind slots. Square panels show each concept's assigned unit; a diagonal means perfect binding, and ``acc.'' is the binding accuracy.}
}
    \label{fig:activation}
    \vspace{-10pt}
\end{figure*}

%% file: tab_fig/fig-frag.tex
\begin{figure}[!t]
    \centering
    \includegraphics[width=0.95\linewidth]{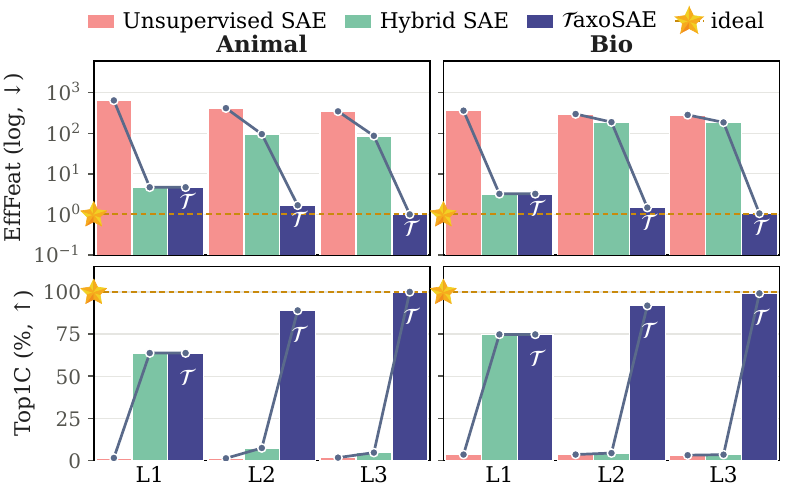}
    \vspace{-5pt}
    \caption{Activation fragmentation (EffFeat) and concentration (Top1C) per taxonomy level on Animal and Bio; dashed lines mark the ideal single unit. \ours{} activations are the most concentrated.}
    \label{fig:frag}
    \vspace{-12pt}
\end{figure}

%% file: tab_fig/tab-swap-bio.tex
\begin{table}[t]
\centering
\normalsize
\definecolor{basegray}{HTML}{F0F0F0}
\definecolor{deltabg}{HTML}{E8F4EC}
\providecommand{\upN}[1]{{\color{gainC}\footnotesize$\uparrow$#1}}
\setlength{\tabcolsep}{2pt}
\renewcommand{\arraystretch}{1.2}
\resizebox{\columnwidth}{!}{%
\begin{tabular}{@{}lcccccc@{}}
\toprule
& \multicolumn{4}{c}{$L_1$ swap}
& \multicolumn{2}{c}{$L_2$ swap} \\
\cmidrule(lr){2-5}\cmidrule(lr){6-7}
& \multicolumn{2}{c}{Answer}
& \multicolumn{2}{c}{Read}
& Answer & Read \\
\cmidrule(lr){2-3}\cmidrule(lr){4-5}\cmidrule(lr){6-6}\cmidrule(lr){7-7}
SAE & Strict & In-set & $L_2$ & $L_3$ & Target-only & $L_3$ \\
\midrule
\rowcolor{basegray}
Random Unsup. & 2.2 & 4.0 & 18.0 & 34.3 & 0.0 & 31.0 \\
DiffMean Unsup.
& \underline{81.8} & \underline{83.0}
& \underline{71.1} & \underline{55.9}
& 12.2 & \underline{69.5} \\
Hybrid
& 84.7 & 88.1 & 66.2 & 44.6
& \underline{16.3} & 65.8 \\
\rowcolor{rowtint}
\textbf{\ours}
& \best{84.7} & \best{88.1}
& \best{100.0} & \best{93.2}
& \best{41.8} & \best{100.0} \\
\rowcolor{deltabg}
$\Delta$(\abb\ $-$ \underline{Best}) 
& \upN{2.9} & \upN{5.1}
& \upN{28.9} & \upN{37.3}
& \upN{25.5} & \upN{30.5} \\
\bottomrule
\end{tabular}}
\vspace{-5pt}
\caption{Bio swaps at each system's best strength. Answer: target-value accuracy; Read: target-hypernym readout. Underlines: best baselines (excluding Hybrid for $L_1$, sharing our $L_1$ SAE); \upN{}: gains over them.
}
\label{tab:swap-bio}
\vspace{-15pt}
\end{table}

%% file: tab_fig/fig-selectivity.tex
\begin{figure}[t]
    \centering
    \includegraphics[width=1\linewidth]{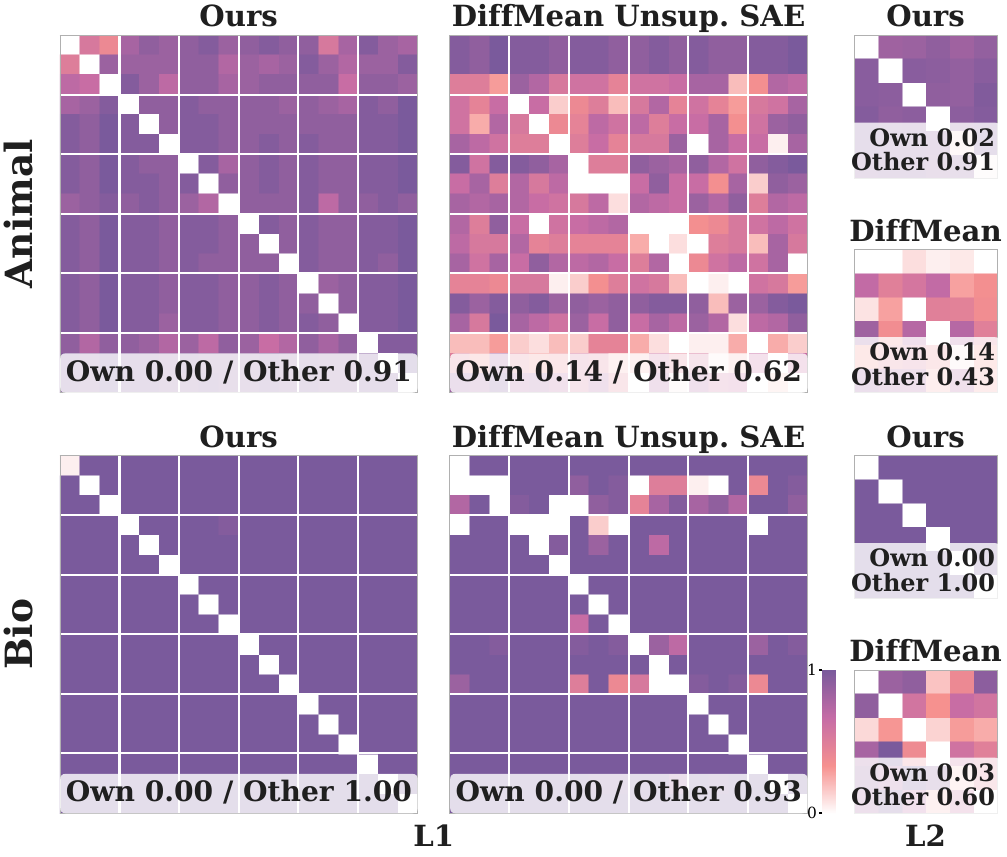}
\vspace{-10pt}
\caption{Selectivity on Animal (top) and Bio (bottom): row $i$ deactivates element $i$'s unit; column $j$ is the exact match on element $j$. Ideally the diagonal is white and the rest dark, as for \ours{}; DiffMean also lightens off-diagonal cells, especially at $L_2$. Own ($\downarrow$)/\edit{Other} ($\uparrow$): mean accuracy on target/non-target elements. \edit{Labeled version: Fig.~\ref{fig:selectivity-labeled}.}}
    \label{fig:selectivity}
    \vspace{-10pt}
\end{figure}

%% file: tab_fig/fig-ablation.tex
\begin{figure}[t]
    \centering
    \includegraphics[width=1\linewidth]{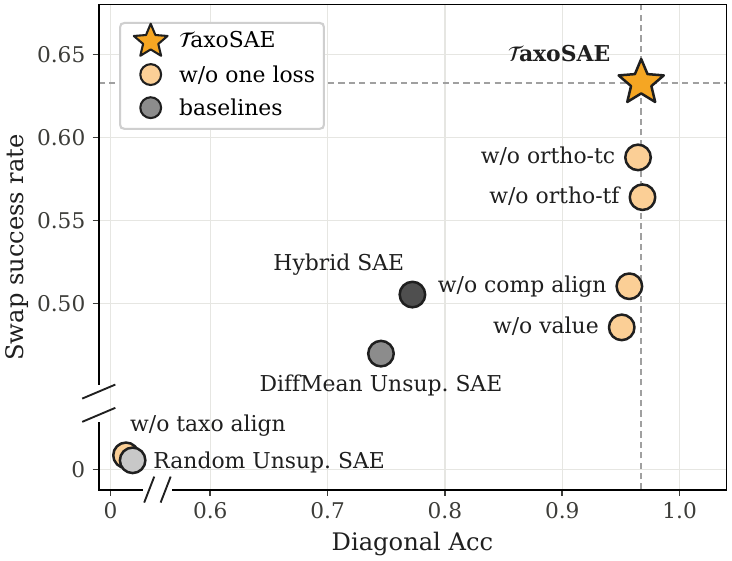}
    \vspace{-15pt}
    \caption{Ablation on Bio (upper right is better). Removing taxonomic alignment collapses both metrics to random; removing others mainly lowers swap success.}
    \label{fig:ablation}
    \vspace{-10pt}
\end{figure}

%% file: sections/conclusion.tex
\section{Conclusion}
\label{sec:conclusion}

We identified that existing SAEs learn a flat feature space, which leaves readout without taxonomic consistency and control without granularity.
We therefore encode taxonomic \texttt{is-a} logic into the feature space and propose \ours{}, which combines stacked SAEs with a taxonomic alignment loss.
On two factual QA taxonomies, \ours{} enables more accurate taxonomy readout and more precise control than the baselines.
Moreover, it reads out unseen concepts at their hypernyms without binding them, which improves data efficiency and suggests that its features align well with the taxonomy.
Future work could extend it to larger models and logical relations beyond taxonomies.

%% file: sections/limitation.tex
\section*{Limitations}
Our evaluation has several limitations.
First, all experiments are in English, and we have not tested whether \ours{} transfers to other languages, where taxonomic knowledge may be expressed and encoded differently.
Second, our taxonomies are relatively simple: \edit{although built from ConceptNet and Freebase, they} have three levels with at most 18 terminal nodes.

%% file: sections/appendix.tex
\section{Datasets}
\label{appendix:datasets}
\input{appendix/dataset}

\section{Results on Qwen2.5-0.5B}
\label{appendix:qwen}
\input{appendix/qwen}

\section{Unseen Terminal Concepts}
\label{appendix:unseen}
\input{appendix/unseen}

\section{\texorpdfstring{\edit{Unseen Phrasings}}{Unseen Phrasings}}
\label{appendix:phrasings}
\input{tab_fig/fig-id-ood-diagonal-acc}
\edit{Beyond evaluating unseen non-terminal nodes, we break down the pooled terminal-question results
($L_1{\to}L_1/L_2/L_3$) in Table~\ref{tab:main}
into in-distribution (ID) evaluation with seen question templates and out-of-distribution (OOD) evaluation with \edit{rephrased questions or synonyms}.
Figure~\ref{fig:acc-id-ood} shows that \ours{} maintains high diagonal accuracy and its advantage over baselines
across datasets under ID and OOD settings. On Animal, OOD accuracy drops with unseen synonyms, but
\ours{} retains a clear lead at non-terminal levels.
On Bio, it maintains near-perfect accuracy under both settings.}

\section{Complement Slot Readout}
\label{appendix:complement}
\input{appendix/complement}

\section{Ablation Study}
\label{appendix:ablation}

\subsection{Loss Ablation}
\input{appendix/ablation-loss}
\label{appendix:ablation-loss}

\subsection{Value head Ablation}
\label{appendix:ablation-value-head}
\input{appendix/ablation-value-head}

\section{Flat AlignSAE Baseline}
\label{appendix:flat}
\input{appendix/flat-alignsae}

\section{Swap Success Results}
\label{appendix:swap-results}
\input{appendix/swap}

\section{\texorpdfstring{\edit{Selectivity Matrices with Concept Labels}}{Selectivity Matrices with Concept Labels}}
\label{appendix:selectivity-labeled}
\begin{figure*}[t]
    \centering
    \includegraphics[width=0.9\textwidth]{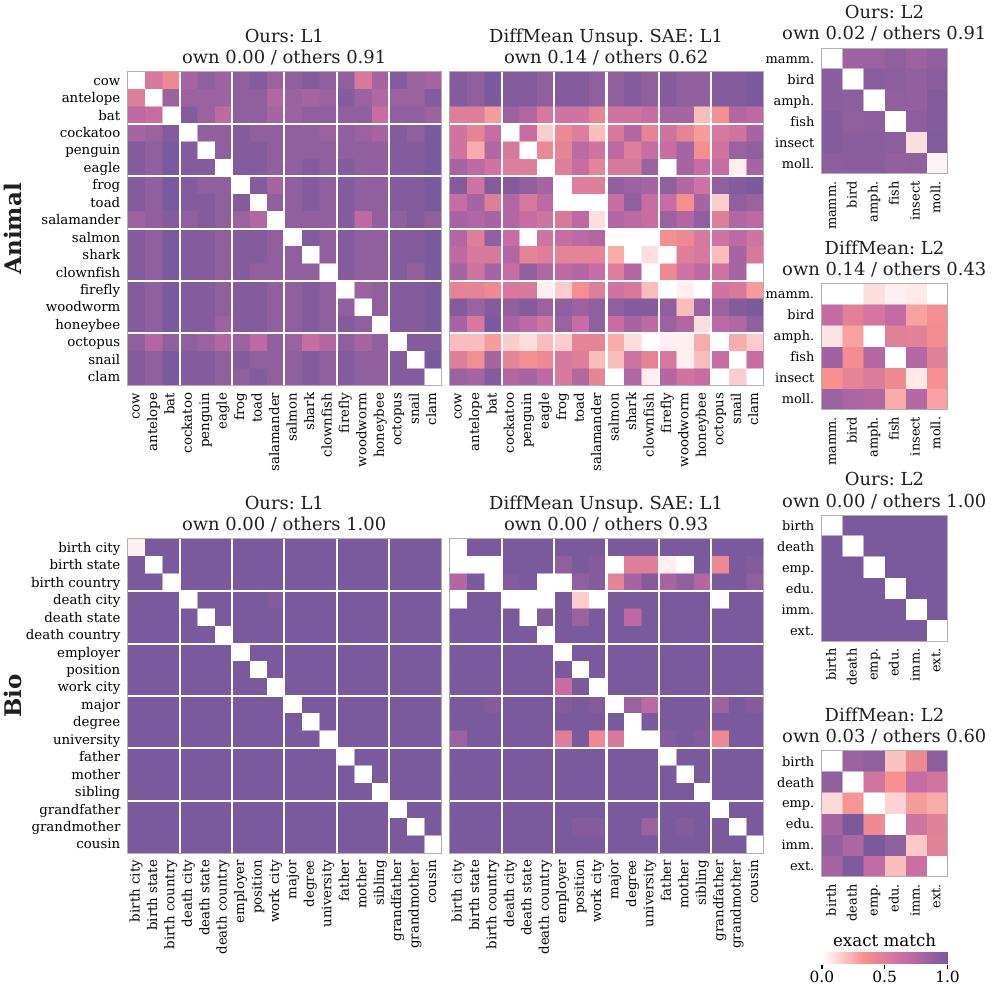}
    \caption{\edit{Fig.~\ref{fig:selectivity} with concept labels. Row $i$ deactivates the unit of element $i$; column $j$ is the exact match on element $j$.}}
    \label{fig:selectivity-labeled}
\end{figure*}

\edit{Figure~\ref{fig:selectivity-labeled} repeats Fig.~\ref{fig:selectivity} with the concept of each row and column.}

\section{Baseline Implementation Details}
\label{appendix:baselines}
\input{appendix/baselines}

\section{Training Details}
\label{appendix:training-details}
\input{appendix/training-details}

\section{Layer-wise Performance}
\label{appendix:layer-wise-performance}
\input{appendix/layer-wise-performance}

%% file: appendix/dataset.tex
\edit{Both taxonomies build on existing resources, while the facts are controlled so that every answer is known,} and both cross a three-level taxonomy with a set of complementary elements (Table~\ref{tab:dataset-stats}).
In \textbf{Bio}, the taxonomy is over biographical relations: 18 terminal relations (e.g., \emph{birth city}, \emph{employer}, \emph{father}) are grouped into 6 intermediate relations (\emph{place of birth}, \emph{place of death}, \emph{employment}, \emph{education}, \emph{immediate family}, \emph{extended family}) and 3 top-level relations (\emph{life places}, \emph{professional background}, \emph{social relations}); \edit{the terminal relations are person relations from Freebase~\citep{bollacker2008freebase}, and we generate the more abstract intermediate and top-level relations above them}; the complementary elements are 20 synthetic people, and each person has one value for each terminal relation.
In \textbf{Animal}, the taxonomy is over animals: 18 terminal animals (e.g., \emph{cow}, \emph{frog}, \emph{octopus}) are grouped into 6 intermediate classes (\emph{mammal}, \emph{bird}, \emph{amphibian}, \emph{fish}, \emph{insect}, \emph{mollusc}) and 2 top-level classes (\emph{vertebrate}, \emph{invertebrate}), \edit{forming a subset of the English \texttt{IsA} relations of ConceptNet~\citep{speer2017conceptnet}}; the complementary elements are 20 properties (e.g., \emph{limb count}, \emph{capable of flight}, \emph{diet type}), and each animal has one value for each property.
A non-terminal node takes, for each complementary element, the set of values of its terminal hyponyms.

\begin{figure*}[h]
\centering
\includegraphics[width=0.95\linewidth]{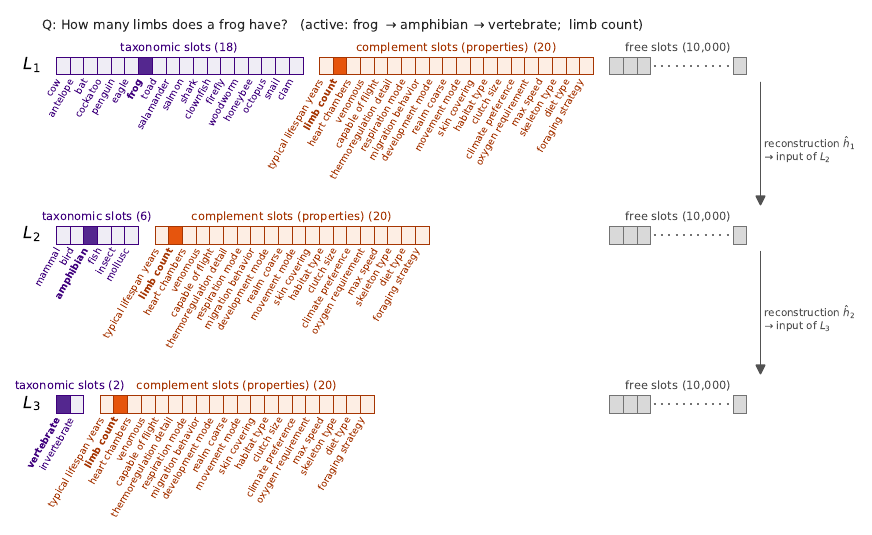}\\[4pt]
\includegraphics[width=0.95\linewidth]{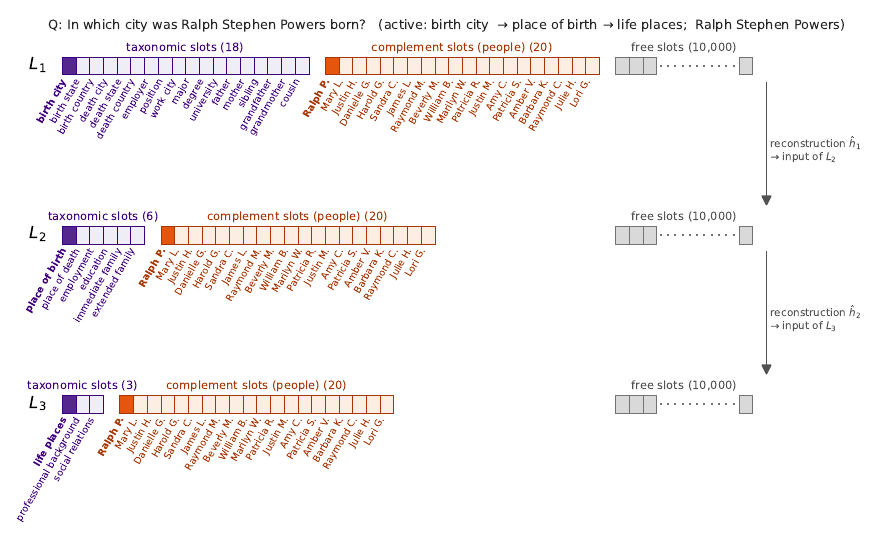}
\caption{Slot layout of the three stacked SAEs on Animal (top) and Bio (bottom). Each level has one taxonomic slot per node at that level, one complement slot per complementary element, and 10{,}000 free slots; $\mathrm{SAE}_{k+1}$ reads the reconstruction of $\mathrm{SAE}_k$. Dark cells mark the slots that the example question should activate: its terminal node and that node's hypernyms in the taxonomic slots, and its complementary element at every level.}
\label{fig:slot-layout}
\end{figure*}

\ours{} is trained only on terminal-level questions.
Bio varies the question template (10 in training) and the surface form of the person's name (4 per person); Animal varies the template (54 in training) and the animal's synonym (5--8 per animal, 108 in total).
The readout tests add held-out templates, names, or synonyms at the terminal level, and intermediate- and top-level questions, none of which appear in training.
Figure~\ref{fig:slot-layout} shows how one question of each dataset maps onto the slots of the three SAEs.

\begin{table*}[h]
\centering
\small
\setlength{\tabcolsep}{4pt}
\begin{tabular}{@{}p{0.2\linewidth}p{0.37\linewidth}p{0.37\linewidth}@{}}
\toprule
 & \textbf{Bio} (relation taxonomy) & \textbf{Animal} (entity taxonomy) \\
\midrule
Nodes at $L_1$ / $L_2$ / $L_3$ & 18 / 6 / 3 & 18 / 6 / 2 \\
Complementary elements & 20 people & 20 properties \\
Terminal facts & $20\times18=360$ & $18\times20=360$ \\
\midrule
SAE training questions & 14{,}400 (10 templates, 80 name forms) & 25{,}014 (54 templates, 108 animal synonyms) \\
Readout, $L_1$ questions & 800 in-distribution / 800 new template / 800 new name / 720 both new & 1{,}080 in-distribution / 720 new template / 720 new synonym / 720 both new \\
Readout, $L_2$ questions & 240 training templates / 240 new templates / 800 extra templates & 600 seen templates / 240 new templates / 600 new synonyms / 240 both new / 1{,}440 extra surface forms \\
Readout, $L_3$ questions & 120 training templates / 120 new templates / 600 extra templates & 200 seen templates / 80 new templates / 200 new synonyms / 80 both new / 720 extra surface forms \\
\midrule
Terminal-level swap pairs & 2{,}550 & 1{,}745 \\
Intermediate-level swap pairs (across top-level nodes) & 600 & 347 \\
\bottomrule
\end{tabular}
\caption{Dataset statistics. Only terminal-level questions are used to train \ours{}; intermediate- and top-level questions appear only in the readout tests.}
\label{tab:dataset-stats}
\end{table*}

%% file: appendix/qwen.tex
To test whether \ours{} transfers beyond GPT-2, we repeat the Bio experiments with Qwen2.5-0.5B (base), a 24-layer model with a hidden dimension of 896 and a modern architecture (RoPE, RMSNorm, SwiGLU).
We fine-tune it on the same data and for the same number of steps as GPT-2, with a learning rate of $10^{-4}$; the fine-tuned model answers all terminal-level factual and hypernymy questions correctly.
We read the middle layer, layer 12, which has the same relative depth as layer 6 in GPT-2. We did not tune this choice, and other layers may give better results.
\ours{} and all baselines use the settings of the main experiments, except that the swap strength $\alpha$ is searched over a wider grid, $\{20, 50, 100, 150, 200,$ $300, 500, 700, 1000, 2000\}$, because the best strength shifts with the base model; each system is reported at its own best $\alpha$. Results are from a single seed.

Table~\ref{tab:qwen} shows that the conclusions of Table~\ref{tab:main} hold.
\ours{} reads non-terminal concepts best, 15.6--33.8 points above the strongest baseline, and represents each non-terminal concept with 1.0--1.3 effective features, against more than 100 for the baselines.
It also gives the best control at both levels, and its slots remove a single concept cleanly: deactivating a terminal or intermediate node drops accuracy on that node's own questions to 0.3\% while keeping at least 99.8\% on the others, whereas DiffMean cannot bring a node's own accuracy below 5\% within the tested strengths.
The absolute swap success is lower than on GPT-2 for all systems (60.5\% and 13.7\% for \ours{}), which may partly reflect the untuned layer.

\begin{table*}[t]
\centering
\small
\definecolor{basegray}{HTML}{F0F0F0}
\definecolor{deltabg}{HTML}{E8F4EC}
\newcommand{\upN}[1]{{\color{gainC}\footnotesize$\uparrow$#1}}
\newcommand{\dnbN}[1]{{\color{gainC}\footnotesize$\downarrow$#1}}
\setlength{\tabcolsep}{2pt}
\newcommand{\dashsep}{\lower1.4\dp\strutbox\hbox{\color{black!22}\vbox to \dimexpr1.4\ht\strutbox+1.4\dp\strutbox\relax{\xleaders\vbox to 2.6pt{\vss\hrule width 0.45pt height 1.4pt}\vfill}}}
\renewcommand{\arraystretch}{1.4}
\resizebox{\textwidth}{!}{%
\begin{tabular}{l*{3}{w{c}{0.99cm}}!{\dashsep}*{2}{w{c}{0.99cm}}!{\dashsep}w{c}{0.99cm}*{3}{w{c}{0.98cm}}*{5}{w{c}{0.84cm}}}
\toprule
& \multicolumn{6}{c}{Diagonal Acc (\%) $\uparrow$} & \multicolumn{3}{c}{EffFeat $\downarrow$} & \multicolumn{3}{c}{Top1C (\%) $\uparrow$} & \multicolumn{2}{c}{Swap (\%) $\uparrow$} \\
\cmidrule(lr){2-7}\cmidrule(lr){8-10}\cmidrule(lr){11-13}\cmidrule(lr){14-15}
SAE & $L_1$${\to}$$L_1$ & $L_1$${\to}$$L_2$ & $L_1$${\to}$$L_3$ & $L_2$${\to}$$L_2$ & $L_2$${\to}$$L_3$ & $L_3$${\to}$$L_3$ & $L_1$ & $L_2$ & $L_3$ & $L_1$ & $L_2$ & $L_3$ & $L_1$ & $L_2$ \\
\midrule
\multicolumn{15}{l}{\emph{Bio relations}, Qwen2.5-0.5B (layer 12)} \\
\rowcolor{basegray}Random Unsup.\ SAE & 1.7 & 0.0 & 0.0 & 0.0 & 0.0 & 0.0 & 138.6 & 135.9 & 132.5 & 6.4 & 7.5 & 6.8 & 1.1 & 0.3 \\
DiffMean Unsup.\ SAE & \second{51.5} & \second{73.5} & 60.6 & 48.9 & \second{56.2} & \second{55.8} & 138.6 & 135.9 & 132.5 & 6.4 & \second{7.5} & 6.8 & \second{44.1} & \second{4.3} \\
Hybrid SAE & \best{100.0} & 63.6 & \second{66.2} & \second{52.1} & 46.5 & 49.2 & \best{3.4} & \second{123.4} & \second{101.3} & \best{71.9} & 4.6 & \second{8.1} & \best{60.5} & 4.0 \\
\rowcolor{rowtint}
\textbf{\ours} & \best{100.0} & \best{100.0} & \best{100.0} & \best{67.7} & \best{88.9} & \best{82.5} & \best{3.4} & \best{1.3} & \best{1.0} & \best{71.9} & \best{95.3} & \best{99.3} & \best{60.5} & \best{13.7} \\
\rowcolor{deltabg}
$\Delta$(\abb\ $-$ Rand.) & \upN{98.3} & \upN{100.0} & \upN{100.0} & \upN{67.7} & \upN{88.9} & \upN{82.5} & \dnbN{135.2} & \dnbN{134.7} & \dnbN{131.5} & \upN{65.5} & \upN{87.7} & \upN{92.5} & \upN{59.5} & \upN{13.4} \\
\bottomrule
\end{tabular}}
\vspace{-5pt}
\caption{Results on Bio with Qwen2.5-0.5B (layer 12), in the format of Table~\ref{tab:main}. Swap: $L_1$ Strict accuracy over all pairs and $L_2$ Target-only coverage@$|U|$ over \edit{pairs from different top-level nodes}, each system at its best $\alpha$. Hybrid SAE shares the $L_1$ module of \ours{}, so its $L_1$ columns are identical. Bold/underline: best/second-best; $\Delta$: \ours{} minus Random Unsup.\ SAE. \edit{The conclusions of Table~\ref{tab:main} hold on Qwen.}}
\label{tab:qwen}
\end{table*}

\edit{\paragraph{Layer Sweep on Qwen.}
To see how the untuned layer choice affects these results, we train \ours{} on each even layer of Qwen2.5-0.5B with the same settings and test readout and swaps over the grid $\{20, 50, 100, 150, 200, 300, 500, 1000\}$ (Table~\ref{tab:layer-sweep-qwen}).
Readout is high at every layer, while control improves with depth up to layers 16--18 ($L_1$ swap 76.2\%, $L_2$ swap 36.5\%), 15.7 and 22.8 points above layer 12, and declines after that, as on GPT-2.}

\begin{table}[t]
\centering
\small
\setlength{\tabcolsep}{4pt}
\renewcommand{\arraystretch}{1.12}
\resizebox{\columnwidth}{!}{%
\begin{tabular}{l ccc cc}
\toprule
& \multicolumn{3}{c}{Diagonal Acc (\%) $\uparrow$} & \multicolumn{2}{c}{Swap (\%) $\uparrow$} \\
\cmidrule(lr){2-4}\cmidrule(lr){5-6}
Layer & $L_1$ & $L_2$ & $L_3$ & $L_1$ & $L_2$ \\
\midrule
$L2$ & 95.1 & \best{76.3} & 74.3 & 3.9 & 0.3 \\
$L4$ & 96.4 & 71.7 & 71.3 & 10.1 & 1.0 \\
$L6$ & \best{100.0} & 74.4 & 69.5 & 26.2 & 0.5 \\
$L8$ & 99.9 & 75.9 & \best{82.5} & 49.6 & 7.5 \\
$L10$ & \best{100.0} & 72.2 & 67.7 & 50.3 & 9.2 \\
\rowcolor{rowtint}$L12$ & \best{100.0} & 67.7 & \best{82.5} & 60.5 & 13.7 \\
$L14$ & \best{100.0} & 66.4 & 65.3 & 56.3 & 16.0 \\
$L16$ & \best{100.0} & 66.3 & 73.3 & \best{76.2} & 26.7 \\
$L18$ & \best{100.0} & 65.6 & 66.3 & 72.1 & \best{36.5} \\
$L20$ & \best{100.0} & 62.8 & 71.8 & 70.8 & 24.2 \\
$L22$ & \best{100.0} & 70.1 & 79.7 & 55.5 & 11.0 \\
\bottomrule
\end{tabular}}
\caption{\edit{Layer-wise performance of \ours{} on Bio with Qwen2.5-0.5B (24 layers), in the format of Table~\ref{tab:layer-sweep-bio}. Shading marks layer 12, used in Table~\ref{tab:qwen}. Readout is stable across layers, and control is best at layers 16--18.}}
\label{tab:layer-sweep-qwen}
\vspace{-10pt}
\end{table}

%% file: appendix/unseen.tex
\edit{A taxonomy should let the upper levels place a concept that no terminal slot was trained on.
We ask the Animal model (layer 9) about 24 animals outside its 18 terminal concepts, four per \edit{intermediate node} (e.g., \emph{pig}, \emph{dog}, \emph{horse}, and \emph{lion} for \emph{mammal}), filling the 100 question templates of a trained terminal concept with the new name, and read the SAEs as in the readout test (Table~\ref{tab:unseen-leaf}).
No terminal slot fits these animals: the highest $L_1$ score halves relative to the trained concepts, and 72.5\% of the questions fall below the 5th percentile of the trained ones.
Because the $L_1$ slots are trained as an 18-way classifier without a ``none'' option, the weaker slot still wins most of the time, usually that of the closest sibling (\emph{pig}$\to$\emph{cow}, \emph{newt}$\to$\emph{salamander}, \emph{tuna}$\to$\emph{salmon}).
The upper levels still recover the animal's place in the taxonomy: the $L_2$ slots of \ours{} read the right \edit{intermediate node} for 63.9\% of the questions and the $L_3$ slots the right top-level node for 76.1\%, against 26.0\% and 56.8\% for DiffMean Unsup.\ SAE.
Hybrid SAE, which shares the terminal-level SAE of \ours{}, is at chance at $L_2$ (16.7\%), so the gain comes from the supervised upper levels rather than from the $L_1$ read; when $L_1$ points to a terminal concept \edit{under another intermediate node}, the $L_2$ slots of \ours{} still read the right \edit{intermediate node} for 31.1\% of the questions, against 1.4\% for Hybrid SAE.
Errors concentrate on a few animals (\emph{sparrow}, \emph{owl}, \emph{goldfish}, \emph{cod}, \emph{ant}, \emph{slug}), while 12 of the 24 animals are classified correctly for at least 99\% of their questions.}

\begin{figure*}[t]
\centering
\includegraphics[width=0.8\textwidth]{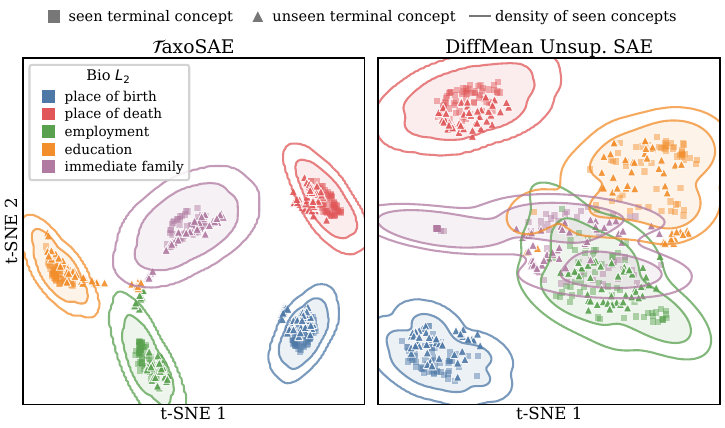}
\caption{\edit{Bio counterpart of Fig.~\ref{fig:unseen-tsne}: t-SNE of the $L_2$ readout scores for seen and unseen terminal relations. The extended-family node is omitted for readability; Table~\ref{tab:unseen-leaf} includes it.}}
\label{fig:unseen-tsne-bio}
\end{figure*}

\edit{On Bio, the terminal concepts are relations, so we add 24 relations outside the 18 trained terminal relations, four under each intermediate relation (e.g., \emph{uncle}, \emph{aunt}, \emph{nephew}, and \emph{niece} under \emph{extended family}), each asked about the 20 people with five question templates.
Table~\ref{tab:unseen-stats} lists all unseen concepts.
The $L_1$ slots again do not fit them (98.0\% of the questions fall below the 5th percentile of the trained ones), while the upper levels of \ours{} read the right intermediate relation for 91.4\% of the questions and the right top-level relation for 94.8\%, against 75.9\% and 69.8\% for the strongest baselines (Table~\ref{tab:unseen-leaf}).
Errors concentrate on \emph{advisor}, which is mostly read as \emph{immediate family} (its questions ask ``who is X's advisor'', like the family relations), and on \emph{uncle}, \emph{nephew}, and \emph{niece}, which are partly read as \emph{immediate family} or abstained on.
Figure~\ref{fig:unseen-tsne-bio} shows the Bio counterpart of Fig.~\ref{fig:unseen-tsne}.}

\begin{table}[t]
\centering
\small
\setlength{\tabcolsep}{3pt}
\renewcommand{\arraystretch}{1.1}
\resizebox{\columnwidth}{!}{%
\begin{tabular}{ll}
\toprule
$L_2$ node & Unseen terminal concepts \\
\midrule
\multicolumn{2}{l}{\emph{Animal} (24 animals, 100 questions each)} \\
mammal & pig, dog, horse, lion \\
bird & sparrow, owl, parrot, pigeon \\
amphibian & newt, axolotl, caecilian, mudpuppy \\
fish & tuna, trout, goldfish, cod \\
insect & ant, butterfly, beetle, mosquito \\
mollusc & squid, oyster, slug, mussel \\
\midrule
\multicolumn{2}{l}{\emph{Bio} (24 relations, 20 people $\times$ 5 templates each)} \\
place of birth & birth hospital, neighborhood, county, continent \\
place of death & death hospital, neighborhood, county, continent \\
employment & department, supervisor, work state, work country \\
education & high school, graduation year, advisor, minor \\
immediate family & spouse, son, daughter, child \\
extended family & uncle, aunt, nephew, niece \\
\bottomrule
\end{tabular}}
\caption{\edit{Unseen terminal concepts, four per intermediate node. None of them appears in SAE training; each unseen concept is compared with 100 questions of each trained terminal concept.}}
\label{tab:unseen-stats}
\vspace{-10pt}
\end{table}

\begin{table}[t]
\centering
\small
\setlength{\tabcolsep}{3pt}
\renewcommand{\arraystretch}{1.2}
\resizebox{\columnwidth}{!}{%
\begin{tabular}{l cc ccc}
\toprule
& \multicolumn{2}{c}{$L_1$ read} & \multicolumn{3}{c}{Upper levels read (\%) $\uparrow$} \\
\cmidrule(lr){2-3}\cmidrule(lr){4-6}
SAE & Top score & Abstain (\%) & $L_2$ & $L_3$ & $L_2$ | $L_1$ off \\
\midrule
\multicolumn{6}{l}{\emph{Animal} (layer 9)} \\
DiffMean Unsup.\ SAE & 0.5$\times$ & 27.6 & 26.0 & 56.8 & 28.2 \\
Hybrid SAE & 0.5$\times$ & 12.9 & 16.7 & 50.5 & 1.4 \\
\rowcolor{rowtint}\textbf{\ours} & 0.5$\times$ & 12.9 & \best{63.9} & \best{76.1} & \best{31.1} \\
Chance & -- & -- & 16.7 & 50.0 & -- \\
\midrule
\multicolumn{6}{l}{\edit{\emph{Bio} (layer 6)}} \\
\edit{DiffMean Unsup.\ SAE} & \edit{0.6$\times$} & \edit{11.7} & \edit{75.9} & \edit{59.3} & \edit{68.1} \\
\edit{Hybrid SAE} & \edit{0.5$\times$} & \edit{10.4} & \edit{54.8} & \edit{69.8} & \edit{26.2} \\
\rowcolor{rowtint}\edit{\textbf{\ours}} & \edit{0.5$\times$} & \edit{10.4} & \edit{\best{91.4}} & \edit{\best{94.8}} & \edit{\best{68.4}} \\
\edit{Chance} & \edit{--} & \edit{--} & \edit{16.7} & \edit{33.3} & \edit{--} \\
\bottomrule
\end{tabular}}
\caption{\edit{Readout of unseen terminal concepts (Table~\ref{tab:unseen-stats}). Top score: median of the highest $L_1$ score, relative to the trained concepts. $L_2$ | $L_1$ off: $L_2$ accuracy on questions whose $L_1$ read falls in another intermediate node. Hybrid SAE shares the $L_1$ module of \ours{}. The upper levels of \ours{} place most unseen concepts under the right intermediate and top-level node on both datasets.}}
\label{tab:unseen-leaf}
\end{table}

%% file: tab_fig/fig-id-ood-diagonal-acc.tex
\begin{figure}[t]
    \centering
    \includegraphics[width=1\linewidth]{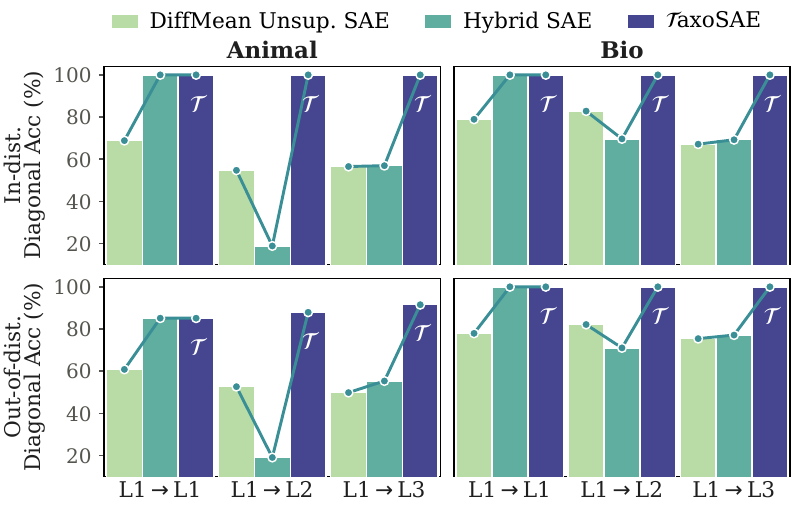}
    \vspace{-10pt}
    \caption{
    Diagonal accuracy on terminal-level questions read at $L_1$--$L_3$, with seen templates (ID, top) and unseen rephrasings or synonyms (OOD, bottom).
    \edit{\ours{} is best at every level in both settings.}
    }
    \label{fig:acc-id-ood}
    \vspace{-15pt}
\end{figure}

%% file: appendix/complement.tex
\edit{Besides the taxonomic slots, every SAE of \ours{} has one complement slot per complementary element: a person in Bio and a property in Animal (Sec.~\ref{sec:method}).
We test whether the complement slots of $\mathrm{SAE}_1$ read out the complementary element of a question, in the same way as Test A: on terminal-level questions, the complement unit with the largest activation must be that of the question's complementary element, with the same 10\% abstention margin and macro accuracy over elements.
For DiffMean, the unit of each element is the feature of the unsupervised terminal-level SAE selected by difference-in-means, and Random Unsup.\ SAE is the mean over four fixed assignments.
Hybrid SAE shares the terminal-level SAE of \ours{}, so its results are identical to ours and are omitted.
Table~\ref{tab:complement-readout} shows that these slots read the complementary element almost perfectly on both datasets, including on question templates, names, and synonyms unseen in training.}

\begin{table}[t]
\centering
\small
\definecolor{basegray}{HTML}{F0F0F0}
\setlength{\tabcolsep}{3pt}
\renewcommand{\arraystretch}{1.15}
\resizebox{\columnwidth}{!}{%
\begin{tabular}{l cccc}
\toprule
SAE & In-dist. & New template & New name/syn. & Both new \\
\midrule
\multicolumn{5}{l}{\emph{Bio} (20 people)} \\
\rowcolor{basegray}Random Unsup.\ SAE & 3.7 & 3.6 & 4.2 & 3.5 \\
DiffMean Unsup.\ SAE & 40.6 & 39.0 & 39.6 & 38.9 \\
\rowcolor{rowtint}\textbf{\ours} & \best{100.0} & \best{100.0} & \best{100.0} & \best{99.3} \\
\midrule
\multicolumn{5}{l}{\emph{Animal} (20 properties)} \\
\rowcolor{basegray}Random Unsup.\ SAE & 4.2 & 4.9 & 4.4 & 4.7 \\
DiffMean Unsup.\ SAE & 84.9 & 83.9 & 87.8 & 84.7 \\
\rowcolor{rowtint}\textbf{\ours} & \best{100.0} & \best{93.8} & \best{100.0} & \best{93.8} \\
\bottomrule
\end{tabular}}
\caption{\edit{Complement readout (Diagonal Acc, \%) on terminal-level questions, by question tier: in-distribution, new templates, new names (Bio) or synonyms (Animal), and both new.}}
\label{tab:complement-readout}
\end{table}

%% file: appendix/ablation-loss.tex
\paragraph{Complete Loss Ablation Results.}
Tables~\ref{tab:ablation} and~\ref{tab:ablation-control}
report the complete results of Tests A--E on Bio.
Each variant is initialized from the same Stage 1 checkpoint
and retrained in Stage 2 with one loss component removed.
Table~\ref{tab:ablation} reports diagonal readout accuracy
and activation concentration across taxonomy levels,
while Table~\ref{tab:ablation-control} reports terminal-
and intermediate-level swapping and selectivity,
with swap results evaluated at each variant's best $\alpha$.
The detailed results reveal trade-offs obscured by aggregate
scores: \edit{removing} taxonomic--free orthogonality improves
several concentration metrics but lowers intermediate-level
swap coverage from 41.8\% to 24.2\%.
These results reinforce that stronger performance on
individual readout or concentration metrics does not
necessarily translate into better steering across levels.

\begin{table*}[t]
\centering
\small
\setlength{\tabcolsep}{2pt}
\renewcommand{\arraystretch}{1.3}
\resizebox{\textwidth}{!}{%
\begin{tabular}{l*{6}{w{c}{0.99cm}}*{3}{w{c}{0.98cm}}*{3}{w{c}{0.98cm}}}
\toprule
& \multicolumn{6}{c}{(A) Diagonal Acc (\%) $\uparrow$}
& \multicolumn{3}{c}{(D) EffFeat $\downarrow$}
& \multicolumn{3}{c}{(D) Top1C (\%) $\uparrow$} \\
\cmidrule(lr){2-7}\cmidrule(lr){8-10}\cmidrule(lr){11-13}
Variant
& $L_1{\to}L_1$ & $L_1{\to}L_2$ & $L_1{\to}L_3$
& $L_2{\to}L_2$ & $L_2{\to}L_3$ & $L_3{\to}L_3$
& $L_1$ & $L_2$ & $L_3$
& $L_1$ & $L_2$ & $L_3$ \\
\midrule
\rowcolor{rowtint}
\textbf{\ours}
& \best{100.0} & \best{100.0} & \best{100.0}
& \best{69.8} & 92.0 & 87.8
& 3.21 & 1.47 & 1.06 & 74.8 & 91.9 & 99.1 \\
\edit{w/o taxonomic alignment}
& 0.0 & 5.4 & 0.0 & 9.8 & 4.0 & 4.7
& 4.32 & 3.11 & 2.40 & 26.8 & 58.6 & 60.5 \\
w/o complement alignment
& \best{100.0} & \best{100.0} & \best{100.0}
& 67.6 & 85.1 & 80.7
& 3.56 & 1.60 & 1.18 & 70.2 & 89.0 & 95.8 \\
w/o value loss
& \best{100.0} & \best{100.0} & \best{100.0}
& 68.1 & 76.9 & 78.3
& 4.02 & 1.70 & 1.28 & 64.9 & 86.5 & 93.5 \\
w/o taxonomic--complement ortho.
& \best{100.0} & \best{100.0} & \best{100.0}
& 69.0 & 88.9 & 88.0
& 2.84 & 1.41 & \best{1.05}
& 79.6 & 93.0 & \best{99.3} \\
w/o taxonomic--free ortho.
& \best{100.0} & \best{100.0} & \best{100.0}
& 67.2 & \best{92.9} & \best{92.3}
& \best{2.16} & \best{1.32} & 1.06
& \best{86.0} & \best{94.7} & 99.0 \\
\bottomrule
\end{tabular}}
\caption{
Loss ablation on Bio (layer 6): taxonomy readout and
activation concentration. Each variant retrains Stage 2
of \ours{} from the same Stage 1 checkpoint with one loss
component removed.
$L_1$/$L_2$/$L_3$ denote terminal/intermediate/top levels.
(A) Diagonal Acc: $L_q{\to}L_r$ denotes readout of level-$r$
taxonomic elements from level-$q$ questions.
(D) EffFeat and Top1C measure activation concentration.
Bold marks the best value in each column.
\edit{Removing taxonomic alignment collapses readout; the other losses change it far less.}
}
\label{tab:ablation}
\end{table*}

\begin{table*}[t]
\centering
\small
\setlength{\tabcolsep}{3pt}
\renewcommand{\arraystretch}{1.3}
\resizebox{\textwidth}{!}{%
\begin{tabular}{l*{3}{c}*{3}{c}*{4}{c}}
\toprule
& \multicolumn{3}{c}{(B) $L_1$ swap}
& \multicolumn{3}{c}{(C) $L_2$ swap}
& \multicolumn{4}{c}{(E) Selectivity (\%)} \\
\cmidrule(lr){2-4}\cmidrule(lr){5-7}\cmidrule(lr){8-11}
Variant
& $\alpha$ & Strict (\%) $\uparrow$ & In-set (\%) $\uparrow$
& $\alpha$ & Cov.\ (\%) $\uparrow$ & Exact (\%) $\uparrow$
& $L_1$ Own $\downarrow$ & $L_1$ Other $\uparrow$
& $L_2$ Own $\downarrow$ & $L_2$ Other $\uparrow$ \\
\midrule
\rowcolor{rowtint}
\textbf{\ours}
& 100 & 84.7 & 88.1 & 100 & 41.8 & 41.8
& 0.0 & \best{100.0} & \best{0.0} & \best{100.0} \\
\edit{w/o taxonomic alignment}
& 200 & 1.3 & 2.4 & 20 & 2.2 & 2.2
& 77.8 & 77.8 & 75.0 & 75.7 \\
w/o complement alignment
& 100 & 60.1 & 63.7 & 100 & \best{42.0} & \best{42.0}
& 0.8 & \best{100.0} & \best{0.0} & \best{100.0} \\
w/o value loss
& 100 & 82.0 & 82.8 & 200 & 15.2 & 15.2
& \best{0.0} & \best{100.0} & 0.8 & 99.6 \\
w/o taxonomic--complement ortho.
& 100 & 80.6 & 86.5 & 100 & 37.0 & 37.0
& 4.7 & \best{100.0} & \best{0.0} & \best{100.0} \\
w/o taxonomic--free ortho.
& 200 & \best{88.6} & \best{90.5} & 200 & 24.2 & 24.2
& \best{0.0} & \best{100.0} & \best{0.0} & \best{100.0} \\
\bottomrule
\end{tabular}}
\caption{
Loss ablation on Bio (layer 6): steering and selectivity
for the same variants as Table~\ref{tab:ablation}.
Each swap test uses the variant's best steering strength $\alpha$.
(B) $L_1$ swap: strict and in-target-set accuracy.
(C) $L_2$ swap on \edit{pairs from different top-level nodes}: Target-only coverage@$|U|$
and Changed exact-set@$|U|$.
(E) Selectivity: Own ($\downarrow$) / Other ($\uparrow$)
exact match when the taxonomic slot corresponding to one
$L_1$ or $L_2$ element is ablated.
Bold marks the best value in each column.
\edit{Removing any loss lowers at least one swap score, while selectivity is preserved except without taxonomic alignment.}
}
\label{tab:ablation-control}
\end{table*}

%% file: appendix/ablation-value-head.tex
\paragraph{Value Head Ablation.}
\edit{We replace the bilinear value head of \S~\ref{sec:method} with the \edit{per-slot} head of AlignSAE~\citep{yang2025alignsae}: one head per taxonomic slot, which reads only the scalar activation of its own slot and is selected by the node of the question.
As in the loss ablation, Stage 2 is retrained from the same Stage 1 checkpoint with all other settings unchanged.
Table~\ref{tab:ablation-value-head} shows that readout, concentration, and selectivity barely change, but intermediate-level swap drops on both datasets (Bio 41.8\% $\to$ 38.3\%; Animal 27.6\% $\to$ 22.1\%).
These drops exceed the variation of \ours{} across three training seeds (standard deviation 1.5 on Bio and 2.4 on Animal), whereas the changes in terminal-level swap stay within that variation.
The bilinear head thus mainly helps edits at the intermediate level, where an edited slot must change the answers of all terminal nodes below it.}

\begin{table}[t]
\centering
\small
\setlength{\tabcolsep}{3pt}
\renewcommand{\arraystretch}{1.15}
\resizebox{\columnwidth}{!}{%
\begin{tabular}{l ccc cc c}
\toprule
& \multicolumn{3}{c}{Diagonal Acc (\%) $\uparrow$} & \multicolumn{2}{c}{Swap (\%) $\uparrow$} & EffFeat $\downarrow$ \\
\cmidrule(lr){2-4}\cmidrule(lr){5-6}\cmidrule(lr){7-7}
Value head & $L_2{\to}L_2$ & $L_2{\to}L_3$ & $L_3{\to}L_3$ & $L_1$ & $L_2$ & $L_2$ \\
\midrule
\multicolumn{7}{l}{\emph{Bio} (layer 6)} \\
\rowcolor{rowtint}Bilinear (\ours{}) & 69.8 & 92.0 & 87.8 & 84.7 & \best{41.8} & 1.47 \\
\edit{Per-slot} & 69.5 & 91.5 & 87.2 & 77.1 & 38.3 & 1.47 \\
\midrule
\multicolumn{7}{l}{\emph{Animal} (layer 9)} \\
\rowcolor{rowtint}Bilinear (\ours{}) & 47.9 & 86.6 & 50.6 & 43.6 & \best{27.6} & 1.68 \\
\edit{Per-slot} & 49.2 & 86.6 & 50.6 & 40.7 & 22.1 & 1.68 \\
\bottomrule
\end{tabular}}
\caption{\edit{Value head ablation. Per-slot: one head per taxonomic slot, reading only that slot, as in AlignSAE. Swap: terminal-level Strict accuracy and intermediate-level Target-only coverage@$|U|$ on \edit{pairs from different top-level nodes}, each at the best $\alpha$. Terminal-level readout ($L_1{\to}L_1/L_2/L_3$) is identical for the two heads on Bio (100\%) and within 0.2 points on Animal.}}
\label{tab:ablation-value-head}
\end{table}

%% file: appendix/flat-alignsae.tex
\begin{figure*}[t]
\centering
\includegraphics[width=0.95\linewidth]{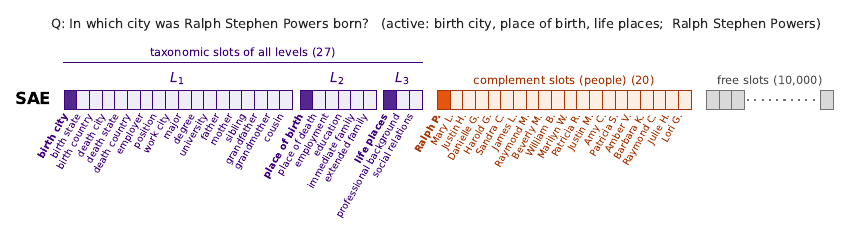}
\caption{\edit{Slot layout of the flat AlignSAE baseline on Bio: a single SAE whose taxonomic slots cover all three levels ($L_1$/$L_2$/$L_3$: 18/6/3), followed by complement and free slots. Dark cells mark the slots that the example question should activate. Compare the stacked layout of \ours{} in Fig.~\ref{fig:slot-layout}.}}
\label{fig:flat-layout}
\end{figure*}

\edit{To test whether the stack is needed, we train a flat AlignSAE~\citep{yang2025alignsae} that binds all levels in a single SAE (Fig.~\ref{fig:flat-layout}).
It reads $h^{(0)}$ directly and has one taxonomic slot for every node of every level, 20 complement slots, and 10{,}000 free slots.
Each level's group of taxonomic slots is supervised with that level's label of the question, derived from the taxonomy, with no stacking and no derived (\texttt{is-a}) targets.
All other losses, weights, and the two-stage schedule (100 + 500 epochs) follow \ours{}, and the tests are run in the same way, editing the slots of the swapped level only.}

\edit{Table~\ref{tab:flat-alignsae} reports the results on Bio (layer 6).
Readout is close to that of \ours{}: both read terminal-level questions perfectly at every level, and the flat SAE reads non-terminal questions about as well.
Control is far weaker.
A terminal swap succeeds for only 5.5\% of the questions, against 84.7\% for \ours{}: editing a terminal slot leaves the intermediate and top-level slots of the same SAE on the source concept, so after a swap across intermediate nodes the flat SAE still reads the source's intermediate node in every case.
In \ours{}, the edited reconstruction feeds the higher SAEs, which read the target's intermediate node in every case.
The intermediate swap shows the same gap (18.3\% vs.\ 41.8\%), and the flat SAE never reads the target's top-level node, since editing its intermediate slot leaves its top-level slots unchanged.
Supervision alone thus suffices for readout, while consistent control across levels requires the stacked structure.}

\begin{table}[t]
\centering
\small
\setlength{\tabcolsep}{3pt}
\renewcommand{\arraystretch}{1.2}
\resizebox{\columnwidth}{!}{%
\begin{tabular}{l ccc cc cc}
\toprule
& \multicolumn{3}{c}{(A) Diagonal Acc (\%) $\uparrow$}
& \multicolumn{2}{c}{(B) $L_1$ swap $\uparrow$}
& \multicolumn{2}{c}{(C) $L_2$ swap $\uparrow$} \\
\cmidrule(lr){2-4}\cmidrule(lr){5-6}\cmidrule(lr){7-8}
SAE & $L_2{\to}L_2$ & $L_2{\to}L_3$ & $L_3{\to}L_3$
& Strict & $L_2$ read & Cov. & $L_3$ read \\
\midrule
Flat AlignSAE & 69.3 & 89.0 & \best{91.7} & 5.5 & 0.0 & 18.3 & 0.0 \\
\rowcolor{rowtint}\textbf{\ours} & \best{69.8} & \best{92.0} & 87.8 & \best{84.7} & \best{100.0} & \best{41.8} & \best{100.0} \\
\bottomrule
\end{tabular}}
\caption{\edit{Flat AlignSAE vs.\ \ours{} on Bio (layer 6). Both read terminal-level questions perfectly at every level ($L_1{\to}L_1/L_2/L_3$, not shown). Swaps use each model's best $\alpha$ (flat: 20 for $L_1$, 50 for $L_2$; \ours{}: 100 for both). $L_2$ read: the SAE reads the target's intermediate node after a terminal swap across intermediate nodes; $L_3$ read: it reads the target's top-level node after an intermediate swap. For the flat SAE, both reads use the edited code itself, which has no higher level to pass the edit to. Readout is similar, but only \ours{} keeps the levels consistent after an edit.}}
\label{tab:flat-alignsae}
\end{table}

%% file: appendix/swap.tex
\paragraph{Evaluation details.}
\edit{In the readout test (Test A), a read counts as wrong when the most active unit is not clearly ahead, i.e., its activation exceeds the second highest by less than 10\%. A swap encodes the hidden state with the SAE of the edited level, sets the unit of the source node to 0 and that of the target node to the strength $\alpha$, and decodes. For a terminal-level swap the edit is made in $\mathrm{SAE}_1$; for an intermediate-level swap the hidden state is passed through $\mathrm{SAE}_1$ and the edit is made in $\mathrm{SAE}_2$. On Bio, swaps follow \S~\ref{section:metrics}: only the last prompt token is edited, and its hidden state is replaced with the decoded state $\mathrm{Dec}(z')$. On Animal, the edit is applied at every prompt position in an error-preserving form that keeps the SAE reconstruction error: each hidden state $h$ becomes $h+\mathrm{Dec}(z')-\mathrm{Dec}(z)$, where $z$ and $z'$ are its activations before and after the edit. Deactivation (Test E) always uses the error-preserving form at the last prompt token. We sweep $\alpha\in\{20,50,100,200,500\}$ on Bio and $\alpha\in\{10,20,50,100,150,200,300,500\}$ on Animal and report each system at its best $\alpha$. For the selectivity test (Test E), the removal strength is swept over $\{1,2,5,10,20,40,80,160\}$, and exact match requires every answer token to be the most likely one under teacher forcing.}

\begin{table*}[t]
\centering
\small
\definecolor{basegray}{HTML}{F0F0F0}
\setlength{\tabcolsep}{4pt}
\renewcommand{\arraystretch}{1.2}
\begin{tabular}{l c cc cc @{\hspace{12pt}} c c c}
\toprule
& \multicolumn{5}{c}{$L_1$ swap (all pairs)} & \multicolumn{3}{c}{$L_2$ swap (\edit{different top-level nodes})} \\
\cmidrule(lr){2-6}\cmidrule(lr){7-9}
& & \multicolumn{2}{c}{Answer (\%) $\uparrow$} & \multicolumn{2}{c}{Read after edit (\%) $\uparrow$} & & Answer (\%) $\uparrow$ & Read after edit (\%) $\uparrow$ \\
\cmidrule(lr){3-4}\cmidrule(lr){5-6}\cmidrule(lr){8-8}\cmidrule(lr){9-9}
SAE & $\alpha$ & Strict & In-set & $L_2$ & $L_3$ & $\alpha$ & Target-only & $L_3$ \\
\midrule
\multicolumn{9}{l}{\emph{Bio} ($L_1$: 2{,}550 pairs; $L_2$: 600 pairs)} \\
\rowcolor{basegray}Random Unsup.\ SAE & -- & 2.2 & 4.0 & 18.0 & 34.3 & -- & 0.0 & 31.0 \\
DiffMean Unsup.\ SAE & 200 & 81.8 & 83.0 & 71.1 & 55.9 & 200 & 12.2 & 69.5 \\
Hybrid SAE & 100 & \best{84.7} & \best{88.1} & 66.2 & 44.6 & 200 & 16.3 & 65.8 \\
\rowcolor{rowtint}\textbf{\ours} & 100 & \best{84.7} & \best{88.1} & \best{100.0} & \best{93.2} & 100 & \best{41.8} & \best{100.0} \\
\midrule
\multicolumn{9}{l}{\emph{Animal} ($L_1$: 1{,}745 pairs; $L_2$: 347 pairs)} \\
\rowcolor{basegray}Random Unsup.\ SAE & -- & 5.3 & 24.4 & 18.8 & 64.6 & -- & 9.4 & 44.5 \\
DiffMean Unsup.\ SAE & 150 & 40.5 & 54.7 & 53.0 & 56.4 & 100 & 25.5 & 69.7 \\
Hybrid SAE & 150 & \best{43.6} & \best{59.3} & 68.4 & 59.3 & 150 & 21.4 & 66.6 \\
\rowcolor{rowtint}\textbf{\ours} & 150 & \best{43.6} & \best{59.3} & \best{100.0} & \best{99.2} & 200 & \best{27.6} & \best{100.0} \\
\bottomrule
\end{tabular}
\caption{Swap results with the levels the SAE reads after the edit, each system at its best edit strength $\alpha$. $L_1$ swap over all source--target pairs: Strict, the answer is the \edit{target terminal node's} value; In-set, any value the \edit{target terminal node} holds; Read after edit, the SAE reads the \edit{target's intermediate ($L_2$) and top-level ($L_3$) hypernyms}. $L_2$ swap on \edit{pairs from different top-level nodes}: Target-only coverage@$|U|$, and whether the SAE reads the \edit{target intermediate node's top-level hypernym} ($L_3$) after the edit. Hybrid SAE shares the $L_1$ module of \ours{}, so its $L_1$ answers are identical, but it reads $L_2$ and $L_3$ with its own unsupervised SAEs; Random Unsup.\ SAE is the mean over four fixed assignments. \edit{After every swap, \ours{} reads the target's hypernyms in 93.2--100\% of cases, against at most 71.1\% for the baselines.}}
\label{tab:swap-pooled}
\end{table*}

\paragraph{Complete Swap Results.}
Table~\ref{tab:swap-pooled} reports answer performance and
taxonomic readout after steering on Bio and Animal,
with each system evaluated at its best steering strength $\alpha$.
Terminal-level swaps cover all source--target pairs,
while intermediate-level swaps use pairs from different \edit{top-level nodes}.
In addition to target-answer metrics, we evaluate whether
the edited activations encode the target element's hypernyms.
After terminal-level swaps, \ours{} achieves 100\%
intermediate-level readout on both datasets and \edit{top-level}
readout of 93.2\% on Bio and 99.2\% on Animal.
After intermediate-level swaps, \edit{top-level} readout reaches
100\% on both datasets.
Although Hybrid SAE shares the same terminal-level answer
performance, its non-terminal readout is substantially weaker.
These results show that steering with \ours{} aligns
higher-level readout with the target element, even when
the generated answer does not always match the target.

%% file: appendix/baselines.tex
\paragraph{Architecture and Training.}
Each unsupervised SAE has 10,000 features, a ReLU encoder,
and a linear decoder:
\begin{align}
    z_\ell &= \operatorname{ReLU}
    (W_\ell^{\mathrm{enc}}x_\ell+b_\ell^{\mathrm{enc}}), \\
    \hat{x}_\ell &= W_\ell^{\mathrm{dec}}z_\ell
    +b_\ell^{\mathrm{dec}},
\end{align}
where $x_1=h$ is the frozen LM activation and
$x_\ell=\hat{x}_{\ell-1}$ for $\ell>1$.
Each unsupervised \edit{SAE} minimizes reconstruction loss
plus an L1 penalty on its activations, with coefficient
$10^{-3}$.
No supervised, taxonomic, or orthogonality losses are
applied to these \edit{SAEs}.

We train for 400 epochs with a batch size of 64 and
a learning rate of $3\times10^{-4}$, decayed to $10^{-6}$
using a cosine schedule.
All baselines use the same training activations as \ours{},
extracted from \edit{transformer layer 6 of GPT-2 for Bio and transformer layer 9 for Animal (App.~\ref{appendix:layer-wise-performance}).}
For Hybrid SAE, these settings apply only to the two
unsupervised \edit{SAEs at the upper taxonomy levels}; the terminal-level SAE is
copied from \ours{} and frozen.
Random Unsup.\ SAE reuses the trained DiffMean dictionaries
without additional training.

\paragraph{Differential Activation Scoring.}
For each taxonomic element $g$, we select a feature from
the SAE at its corresponding level.
Let $\mathcal{D}_g$ denote the terminal-level training
questions associated with $g$, and define
\begin{align}
    \mu_{g,f}
    &= \frac{1}{|\mathcal{D}_g|}
    \sum_{x\in\mathcal{D}_g} z_{\ell(g),f}(x), \\
    s(g,f)
    &= \mu_{g,f}
    - \max_{\substack{g'\neq g\\\ell(g')=\ell(g)}}\mu_{g',f}.
\end{align}
Here, $\ell(g)$ is the taxonomy level of $g$.
The selected feature is
$f_g=\arg\max_f s(g,f)$.
This max-margin variant of difference-in-means favors
features that respond more strongly to $g$ than to any
other element at the same level.

For non-terminal elements, labels are derived from the
terminal labels using the known taxonomy.
No non-terminal questions are used for feature selection.
Thus, SAE dictionary training is unsupervised, while
post-hoc feature selection uses labels.
Hybrid SAE applies the same procedure only at the
intermediate and top levels.

\paragraph{Fixed Feature Assignment.}
For Random Unsup.\ SAE, we evaluate four assignments:
one assigns the $i$th taxonomic element to the $i$th feature
at its level, and three randomly assign distinct features
using seeds 1--3.
We report the mean across these four assignments.
Since Random and DiffMean Unsup.\ SAE share the same
dictionaries, their EffFeat and Top1C scores are identical;
these metrics do not depend on feature assignment.

\paragraph{Feature Budget.}
All baselines use one feature per taxonomic element
($k=1$) during evaluation, matching the single-slot budget
of \ours{}.
Although the five highest-scoring features are saved
during feature selection, only the highest-scoring
feature is used for readout and intervention.

%% file: appendix/training-details.tex
\subsection{Base Model Training}
\label{appendix:base-model-training}

\paragraph{Training Data.}
For each dataset, the training corpus combines terminal-level
factual QA, hypernymy QA, and plain-text descriptions of
terminal-level facts.
Bio contains 14,400 factual QA examples, 159 hypernymy QA
examples, and 20 biographies; Animal contains 25,014 factual
QA examples, 1,126 hypernymy QA examples, and 38 descriptions \edit{(Tables~\ref{tab:base-lm-data-bio} and~\ref{tab:base-lm-data-animal})}.
\edit{Hypernymy QA maps terminal nodes to their intermediate and top-level hypernyms and intermediate nodes to top-level ones,
including yes/no checks, and enumerates the hyponyms of
intermediate or top-level nodes.}
These questions provide taxonomic structure without supplying
factual answers for non-terminal nodes.
No non-terminal factual QA is included, so answering such
questions requires generalizing beyond the directly
provided terminal-level facts.

\begin{table}[h]
\centering
\small
\setlength{\tabcolsep}{8pt}
\renewcommand{\arraystretch}{1.12}
\begin{tabular}{lr}
\toprule
\rowcolor{rowtint}
\textbf{Bio: training data} & \textbf{\#} \\
\midrule
Terminal-level factual QA & 14{,}400 \\
Hypernymy QA & 159 \\
\quad terminal${\to}$intermediate & 90 \\
\quad terminal${\to}$top-level & 36 \\
\quad intermediate${\to}$top-level & 24 \\
\quad hyponym enumeration & 9 \\
Non-terminal factual QA & 0 \\
Biographies & 20 \\
\bottomrule
\end{tabular}
\caption{Base-model training data for Bio. Mapping questions
include yes/no checks (e.g., ``Is `birth city' a place of birth
fact?''); hyponym enumeration asks for the members of an
intermediate or top-level node. Biographies are plain-text
descriptions of terminal-level facts.}
\label{tab:base-lm-data-bio}
\end{table}

\begin{table}[h]
\centering
\small
\setlength{\tabcolsep}{8pt}
\renewcommand{\arraystretch}{1.12}
\begin{tabular}{lr}
\toprule
\rowcolor{rowtint}
\textbf{Animal: training data} & \textbf{\#} \\
\midrule
Terminal-level factual QA & 25{,}014 \\
Hypernymy QA & 1{,}126 \\
\quad terminal${\to}$intermediate & 864 \\
\quad terminal${\to}$top-level & 162 \\
\quad intermediate${\to}$top-level & 66 \\
\quad hyponym enumeration & 34 \\
Non-terminal factual QA & 0 \\
Descriptions & 38 \\
\bottomrule
\end{tabular}
\caption{Base-model training data for Animal. Mapping questions
include yes/no checks (e.g., ``Is a cow a mammal?''); hyponym
enumeration asks for the members of an intermediate or top-level
node (e.g., ``Which animals are mammals?''). Descriptions state
terminal-level facts in plain text.}
\label{tab:base-lm-data-animal}
\end{table}

\paragraph{Training Configuration.}
We fine-tune GPT-2 small separately on Bio and Animal
using AdamW with $\epsilon=10^{-6}$ \edit{(Tables~\ref{tab:base-lm-training-bio} and~\ref{tab:base-lm-training-animal})}.
For Bio, we train for 13k steps with a batch size of 64,
using only biographies for the first 1k steps and a
15\% biography share thereafter.
The learning rate is $5\times10^{-4}$, with 500 warm-up
steps and cosine decay to $5\times10^{-5}$.
For Animal, descriptions are mixed with QA data throughout
30k training steps, with a batch size of 32, a learning
rate of $2\times10^{-5}$, and 1,000 warm-up steps.
The loss is computed on answer tokens for QA and on all
tokens for biographies and descriptions.
\edit{With these per-dataset settings, each fine-tuned model answers all of its training factual and hypernymy questions correctly.}
After fine-tuning, we freeze the models and extract
\edit{hidden states} at the final prompt token
from layer 6 for Bio and layer 9 for Animal \edit{(App.~\ref{appendix:layer-wise-performance})}.

\begin{table}[h]
\centering
\small
\setlength{\tabcolsep}{8pt}
\renewcommand{\arraystretch}{1.12}
\begin{tabular}{ll}
\toprule
\rowcolor{rowtint}
\textbf{Bio: fine-tuning} & \\
\midrule
Base model & GPT-2 small \\
Training steps & 13k \\
\quad biographies only & first 1k \\
\quad biography share afterwards & 15\% \\
Batch size & 64 \\
Optimizer & AdamW ($\epsilon{=}10^{-6}$) \\
Learning rate & $5{\times}10^{-4}$ \\
LR schedule & cosine to $5{\times}10^{-5}$ \\
Warm-up steps & 500 \\
Weight decay & 0.1 \\
Max sequence length & 384 \\
Precision & bf16 \\
Seed & 0 \\
Activation layer & 6 \\
\bottomrule
\end{tabular}
\caption{Base-model fine-tuning for Bio. GPT-2 small has
12 transformer layers and a hidden dimension of 768.
The loss is computed on answer tokens for QA and on all
tokens for biographies. \edit{Activations are the hidden state} at the final prompt token.}
\label{tab:base-lm-training-bio}
\end{table}

\begin{table}[h]
\centering
\small
\setlength{\tabcolsep}{8pt}
\renewcommand{\arraystretch}{1.12}
\begin{tabular}{ll}
\toprule
\rowcolor{rowtint}
\textbf{Animal: fine-tuning} & \\
\midrule
Base model & GPT-2 small \\
Training steps & 30k \\
\quad descriptions & mixed into the data \\
Batch size & 32 \\
Optimizer & AdamW ($\epsilon{=}10^{-6}$) \\
Learning rate & $2{\times}10^{-5}$ \\
LR schedule & cosine to $10^{-4}$ \\
Warm-up steps & 1{,}000 \\
Weight decay & 0.01 \\
Max sequence length & 512 \\
Precision & fp32 \\
Activation layer & 9 \\
\bottomrule
\end{tabular}
\caption{Base-model fine-tuning for Animal. GPT-2 small has
12 transformer layers and a hidden dimension of 768.
The loss is computed on answer tokens for QA and on all
tokens for descriptions. \edit{Activations are the hidden state} at the final prompt token.}
\label{tab:base-lm-training-animal}
\end{table}

\subsection{\ours{} Training}
\label{appendix:ours-training}

\edit{Each of the three SAEs of \ours{} has 10,000 free slots, one taxonomic slot per node at its taxonomy level (18/6/3 for Bio and 18/6/2 for Animal), and 20 complement slots, i.e., people in Bio or properties in Animal. The bilinear value head uses a 64-dimensional taxonomy embedding and 128-dimensional projections.}

\begin{table}[t]
\centering
\small
\setlength{\tabcolsep}{3pt}
\begin{tabular}{@{}lcc@{}}
\toprule
\rowcolor{rowtint}
\textbf{Loss term} & \textbf{Stage 1} & \textbf{Stage 2} \\
\midrule
Reconstruction & 1 & 1 \\
Sparsity, free slots & $10^{-3}$ & $10^{-3}$ \\
Sparsity, taxonomic slots & $10^{-3}$ & $10^{-3}$ \\
Sparsity, complement slots & $10^{-3}$ & $10^{-3}$ \\
Taxonomic alignment, terminal (CE) & 0 & 1.0 \\
Taxonomic alignment, non-terminal (CE) & 0 & 5.0 \\
Complement alignment & 0 & 1.0 \\
Value loss & 0 & 1.0 \\
Taxonomic--complement orthogonality & 0 & 0.5 \\
Taxonomic--free orthogonality & 0 & $10^{-2}$ \\
\bottomrule
\end{tabular}
\caption{Loss weights of \ours{}. Both datasets
(\textbf{Bio} and \textbf{Animal}) use the same weights.
Stage~2 is initialized from the Stage~1 checkpoint.}
\label{tab:sae_hparams}
\end{table}

Table~\ref{tab:sae_hparams} lists the loss weights, shared
across Bio and Animal.
Stage 1 runs for 100 epochs using only reconstruction
and sparsity losses, with all other loss weights set to zero.
Stage 2 starts from the Stage 1 checkpoint and trains
for 500 epochs with all losses enabled.
\edit{Both stages use AdamW, a learning rate of $3\times10^{-4}$ with cosine decay to $10^{-6}$, a batch size of 64, gradient clipping at 1.0, and seed 0, on the activations of the terminal-level training questions (14,400 for Bio; 25,014 for Animal).}
The reconstruction weight is 1.0 in both stages.
L1 sparsity is computed separately for free, taxonomic,
and complement slots, each with a weight of $10^{-3}$.

In Stage 2, the taxonomic alignment loss
$\mathcal{L}_{\text{taxo}}$ uses cross-entropy with a weight
of 1.0 at the terminal level and 5.0 at the intermediate
and top levels.
Only terminal-level taxonomic alignment uses ground-truth
labels; non-terminal targets are derived from terminal-level
predictions through the taxonomy.
The complement alignment loss $\mathcal{L}_{\text{comp}}$
and the value loss each receive a weight of 1.0.
Taxonomic--complement orthogonality
$\mathcal{L}_{\perp}^{\text{tc}}$ has a weight of 0.5,
while taxonomic--free orthogonality
$\mathcal{L}_{\perp}^{\text{tf}}$ has a weight of $10^{-2}$.
Following AlignSAE, the latter is computed as the sum of
squared cross-covariances over all taxonomic--free slot pairs.

%% file: appendix/layer-wise-performance.tex
\begin{table}[t]
\centering
\small
\setlength{\tabcolsep}{4pt}
\renewcommand{\arraystretch}{1.12}
\resizebox{\columnwidth}{!}{%
\begin{tabular}{l ccc cc c}
\toprule
& \multicolumn{3}{c}{Diagonal Acc (\%) $\uparrow$} & \multicolumn{2}{c}{Swap (\%) $\uparrow$} & $\downarrow$ \\
\cmidrule(lr){2-4}\cmidrule(lr){5-6}\cmidrule(lr){7-7}
Layer & $L_1$ & $L_2$ & $L_3$ & $L_1$ & $L_2$ & Recon NMSE \\
\midrule
$L0$ & 5.3 & 21.3 & 35.5 & 2.9 & 0.3 & 0.372 \\
$L1$ & 90.1 & 72.1 & \best{98.0} & 0.9 & 0.0 & 0.310 \\
$L2$ & \best{100.0} & 69.0 & 79.8 & 11.4 & 0.3 & 0.252 \\
$L3$ & 99.6 & 70.4 & 85.8 & 47.5 & 14.0 & 0.129 \\
$L4$ & \best{100.0} & 71.8 & 89.0 & 65.5 & 35.2 & 0.082 \\
$L5$ & \best{100.0} & 71.6 & 89.3 & 79.1 & 39.5 & 0.058 \\
\rowcolor{rowtint}$L6$ & \best{100.0} & 69.8 & 87.8 & \best{84.7} & 41.8 & 0.045 \\
$L7$ & \best{100.0} & 73.9 & 89.3 & 65.3 & \best{47.0} & 0.040 \\
$L8$ & \best{100.0} & \best{76.7} & 89.0 & 72.6 & 39.3 & 0.038 \\
$L9$ & \best{100.0} & 76.2 & 92.8 & 63.8 & 28.7 & 0.048 \\
$L10$ & \best{100.0} & 76.6 & 91.0 & 61.2 & 23.7 & 0.092 \\
$L11$ & \best{100.0} & 76.5 & 91.3 & 61.3 & 24.0 & 0.141 \\
$L12$ & \best{100.0} & 76.0 & 92.5 & 53.5 & 7.7 & \best{0.024} \\
\bottomrule
\end{tabular}}
\caption{Layer-wise performance of \ours{} on Bio.
Layer 0 is the embedding output. Bold marks column-best
results; shading marks layer 6, which achieves the highest
mean swap score. \edit{Steering emerges only from the middle layers.}}
\label{tab:layer-sweep-bio}
\end{table}

\paragraph{Layer-Wise Evaluation on Bio.}
Table~\ref{tab:layer-sweep-bio} reports results for \ours{}
trained separately on activations from the embedding output
and each GPT-2 block.
Diagonal accuracy measures same-level taxonomic readout
($L_k{\to}L_k$), using pooled ID and OOD terminal-level questions
and unseen intermediate- and top-level questions.
Swap scores measure terminal-level strict accuracy over
all pairs and intermediate-level target-only coverage@$|U|$
over \edit{pairs from different top-level nodes}, each at its best steering strength
$\alpha$.
Reconstruction NMSE is the terminal-level SAE's reconstruction
error on training activations divided by activation variance.

Terminal-level readout reaches near-perfect accuracy
in early layers, but strong steering emerges later.
Terminal-level swapping peaks at layer 6 (84.7\%),
while intermediate-level swapping peaks at layer 7 (47.0\%).
\edit{Layer 6 gives the best terminal-level swap (84.7\%), the criterion used in the main experiments. All systems use the same layer, and only terminal-level metrics were used for the choice; non-terminal and unseen-concept results were not.}

\begin{table}[t]
\centering
\small
\setlength{\tabcolsep}{4pt}
\renewcommand{\arraystretch}{1.12}
\resizebox{\columnwidth}{!}{%
\begin{tabular}{l ccc cc c}
\toprule
& \multicolumn{3}{c}{Diagonal Acc (\%) $\uparrow$} & \multicolumn{2}{c}{Swap (\%) $\uparrow$} & $\downarrow$ \\
\cmidrule(lr){2-4}\cmidrule(lr){5-6}\cmidrule(lr){7-7}
Layer & $L_1$ & $L_2$ & $L_3$ & $L_1$ & $L_2$ & Recon NMSE \\
\midrule
$L0$ & 3.8 & 17.9 & 50.0 & 5.1 & 5.2 & 0.967 \\
$L1$ & 69.0 & 27.1 & 47.1 & 4.3 & 8.6 & 2.605 \\
$L2$ & 76.4 & 46.2 & 50.4 & 4.1 & 6.2 & 1.775 \\
$L3$ & 76.2 & 41.7 & 51.1 & 5.0 & 7.9 & 0.879 \\
$L4$ & 75.3 & 36.7 & 50.7 & 6.0 & 7.6 & 0.824 \\
$L5$ & 74.7 & 45.4 & 49.4 & 5.6 & 7.6 & 0.480 \\
$L6$ & 75.4 & 46.2 & 53.8 & 11.0 & 10.3 & 0.204 \\
$L7$ & 78.1 & 50.8 & \best{55.4} & 15.0 & 13.1 & 0.144 \\
$L8$ & 78.6 & 51.3 & 51.4 & 21.6 & 11.4 & 0.076 \\
\rowcolor{rowtint}$L9$ & \best{84.3} & 47.9 & 50.6 & \best{43.6} & 27.6 & 0.086 \\
$L10$ & 82.6 & 91.2 & 49.9 & 38.8 & \best{54.5} & 0.115 \\
$L11$ & 81.9 & 95.0 & 50.0 & 37.4 & 34.8 & 0.167 \\
$L12$ & 82.1 & \best{98.8} & 50.3 & 39.8 & 30.3 & \best{0.048} \\
\bottomrule
\end{tabular}}
\caption{\edit{Layer-wise performance of \ours{} on Animal, in the format of Table~\ref{tab:layer-sweep-bio}.
Layer 0 is the embedding output. Bold marks column-best results; shading marks layer 9, used in the main experiments. Steering emerges only from the middle layers.}}
\label{tab:layer-sweep-animal}
\end{table}

\paragraph{\edit{Layer-Wise Evaluation on Animal.}}
\edit{Table~\ref{tab:layer-sweep-animal} repeats this evaluation on Animal.
As on Bio, early layers support little steering: both swap scores stay below 22\% up to layer 8.
Layer 9, used in the main experiments, gives the best terminal-level readout (84.3\%) and terminal-level swap (43.6\%).
Later layers read intermediate concepts better, and layer 10 reaches the highest intermediate-level swap (54.5\%, against 27.6\% at layer 9) and the highest mean swap score.
Top-level readout stays near 50\% at every layer, i.e., chance for the two top-level nodes.}

\section{The Usage of AI Tools}

In this work, large language models were used solely for auxiliary purposes, primarily to refine the language of the manuscript, including grammar correction, improving clarity, and replacing informal expressions with academic phrasing. Importantly, LLMs were not used for research ideation, hypothesis formulation, or experimental design. All conceptual contributions, experimental designs, and method  design were independently developed and carried out by the authors.

%% file: main.bib
@article{bricken2023towards,
  title={Towards monosemanticity: Decomposing language models with dictionary learning},
  author={Bricken, Trenton and Templeton, Adly and Batson, Joshua and Chen, Brian and Jermyn, Adam and Conerly, Tom and Turner, Nick and Anil, Cem and Denison, Carson and Askell, Amanda and others},
  journal={Transformer Circuits Thread},
  volume={2},
  number={5},
  pages={6},
  year={2023}
}

@article{yang2025alignsae,
  title={Alignsae: Concept-aligned sparse autoencoders},
  author={Yang, Minglai and Guo, Xinyu and Shi, Zhengliang and Bi, Jinhe and Bethard, Steven and Surdeanu, Mihai and Pan, Liangming},
  journal={arXiv preprint arXiv:2512.02004},
  year={2025}
}

@misc{hu2020harnessingdeepneuralnetworks,
      title={Harnessing Deep Neural Networks with Logic Rules}, 
      author={Zhiting Hu and Xuezhe Ma and Zhengzhong Liu and Eduard Hovy and Eric Xing},
      year={2020},
      eprint={1603.06318},
      archivePrefix={arXiv},
      primaryClass={cs.LG},
      url={https://arxiv.org/abs/1603.06318}, 
}

@article{lindsey2025biology,
  title={On the biology of a large language model. Transformer Circuits Thread (2025)},
  author={Lindsey, Jack and Gurnee, Wes and Ameisen, Emmanuel and Chen, Brian and Pearce, Adam and Turner, Nicholas L and Citro, Craig and Abrahams, David and Carter, Shan and Hosmer, Basil and others},
  journal={URL: https://transformer-circuits. pub/2025/attribution-graphs/biology. html},
  year={2025}
}

@misc{afzal2025knowingsayingllmrepresentations,
      title={Knowing Before Saying: LLM Representations Encode Information About Chain-of-Thought Success Before Completion}, 
      author={Anum Afzal and Florian Matthes and Gal Chechik and Yftah Ziser},
      year={2025},
      eprint={2505.24362},
      archivePrefix={arXiv},
      primaryClass={cs.CL},
      url={https://arxiv.org/abs/2505.24362}, 
}

@misc{muchane2025hierarchical,
  title={Incorporating Hierarchical Semantics in Sparse Autoencoder Architectures},
  author={Muchane, Mark and Richardson, Sean and Park, Kiho and Veitch, Victor},
  year={2025},
  eprint={2506.01197},
  archivePrefix={arXiv},
  primaryClass={cs.LG}
}

@inproceedings{park2025geometry,
  title={The Geometry of Categorical and Hierarchical Concepts in Large Language Models},
  author={Park, Kiho and Choe, Yo Joong and Jiang, Yibo and Veitch, Victor},
  booktitle={The Thirteenth International Conference on Learning Representations},
  year={2025}
}

@article{belinkov2022probing,
  title={Probing Classifiers: Promises, Shortcomings, and Advances},
  author={Belinkov, Yonatan},
  journal={Computational Linguistics},
  volume={48},
  number={1},
  pages={207--219},
  year={2022}
}

@inproceedings{gao2025scaling,
  title={Scaling and evaluating sparse autoencoders},
  author={Gao, Leo and Dupre la Tour, Tom and Tillman, Henk and Goh, Gabriel and Troll, Rajan and Radford, Alec and Sutskever, Ilya and Leike, Jan and Wu, Jeffrey},
  booktitle={International Conference on Learning Representations},
  volume={2025},
  pages={26721--26754},
  year={2025}
}

@inproceedings{yeo2025understanding,
  title={Understanding Refusal in Language Models with Sparse Autoencoders.},
  author={Yeo, Wei Jie and Prakash, Nirmalendu and Neo, Clement and Satapathy, Ranjan and Lee, Roy Ka-Wei and Cambria, Erik},
  booktitle={EMNLP (Findings)},
  pages={6377--6399},
  year={2025}
}

@article{elhage2022toy,
  title={Toy models of superposition},
  author={Elhage, Nelson and Hume, Tristan and Olsson, Catherine and Schiefer, Nicholas and Henighan, Tom and Kravec, Shauna and Hatfield-Dodds, Zac and Lasenby, Robert and Drain, Dawn and Chen, Carol and others},
  journal={arXiv preprint arXiv:2209.10652},
  year={2022}
}

@inproceedings{huben2024sparse,
  title={Sparse autoencoders find highly interpretable features in language models},
  author={Huben, Robert and Cunningham, Hoagy and Smith, Logan and Ewart, Aidan and Sharkey, Lee},
  booktitle={International Conference on Learning Representations},
  volume={2024},
  pages={7827--7845},
  year={2024}
}

@article{rajamanoharan2024jumping,
  title={Jumping ahead: Improving reconstruction fidelity with jumprelu sparse autoencoders},
  author={Rajamanoharan, Senthooran and Lieberum, Tom and Sonnerat, Nicolas and Conmy, Arthur and Varma, Vikrant and Kram{\'a}r, J{\'a}nos and Nanda, Neel},
  journal={arXiv preprint arXiv:2407.14435},
  year={2024}
}

@inproceedings{chanin2025absorption,
  title={A is for Absorption: Studying Feature Splitting and Absorption in Sparse Autoencoders},
  author={Chanin, David and Wilken-Smith, James and Dulka, Tom{\'a}{\v{s}} and Bhatnagar, Hardik and Golechha, Satvik and Bloom, Joseph},
  booktitle={Advances in Neural Information Processing Systems},
  year={2025},
  note={arXiv:2409.14507}
}

@inproceedings{makelov2025towards,
  title={Towards principled evaluations of sparse autoencoders for interpretability and control},
  author={Makelov, Aleksandar and Lange, Georg and Nanda, Neel},
  booktitle={International Conference on Learning Representations},
  volume={2025},
  pages={33588--33636},
  year={2025}
}

@article{karvonen2025saebench,
  title={SAEBench: A comprehensive benchmark for sparse autoencoders in language model interpretability},
  author={Karvonen, Adam and Rager, Can and Lin, Johnny and Tigges, Curt and Bloom, Joseph and Chanin, David and Lau, Yeu-Tong and Farrell, Eoin and McDougall, Callum and Ayonrinde, Kola and others},
  journal={arXiv preprint arXiv:2503.09532},
  year={2025}
}

@inproceedings{koh2020concept,
  title={Concept bottleneck models},
  author={Koh, Pang Wei and Nguyen, Thao and Tang, Yew Siang and Mussmann, Stephen and Pierson, Emma and Kim, Been and Liang, Percy},
  booktitle={International conference on machine learning},
  pages={5338--5348},
  year={2020},
  organization={Pmlr}
}

@misc{yang2023injectinglogicalconstraintsneural,
      title={Injecting Logical Constraints into Neural Networks via Straight-Through Estimators}, 
      author={Zhun Yang and Joohyung Lee and Chiyoun Park},
      year={2023},
      eprint={2307.04347},
      archivePrefix={arXiv},
      primaryClass={cs.AI},
      url={https://arxiv.org/abs/2307.04347}, 
}

@inproceedings{sun2024head,
  title={Head-to-tail: How knowledgeable are large language models (LLMs)? AKA will LLMs replace knowledge graphs?},
  author={Sun, Kai and Xu, Yifan and Zha, Hanwen and Liu, Yue and Dong, Xin Luna},
  booktitle={Proceedings of the 2024 conference of the North American chapter of the association for computational linguistics: human language technologies (volume 1: long papers)},
  pages={311--325},
  year={2024}
}

@inproceedings{cohen2023crawling,
  title={Crawling the internal knowledge-base of language models},
  author={Cohen, Roi and Geva, Mor and Berant, Jonathan and Globerson, Amir},
  booktitle={Findings of the Association for Computational Linguistics: EACL 2023},
  pages={1856--1869},
  year={2023}
}

@inproceedings{hu2025enabling,
  title={Enabling LLM knowledge analysis via extensive materialization},
  author={Hu, Yujia and Nguyen, Tuan-Phong and Ghosh, Shrestha and Razniewski, Simon},
  booktitle={Proceedings of the 63rd Annual Meeting of the Association for Computational Linguistics (Volume 1: Long Papers)},
  pages={16189--16202},
  year={2025}
}

@article{yu2022generate,
  title={Generate rather than retrieve: Large language models are strong context generators},
  author={Yu, Wenhao and Iter, Dan and Wang, Shuohang and Xu, Yichong and Ju, Mingxuan and Sanyal, Soumya and Zhu, Chenguang and Zeng, Michael and Jiang, Meng},
  journal={arXiv preprint arXiv:2209.10063},
  year={2022}
}

@article{yifei2026researchqa,
  title={Researchqa: Evaluating scholarly question answering at scale across 75 fields with survey-mined questions and rubrics},
  author={Yifei, Li S and Chang, Allen and Malaviya, Chaitanya and Yatskar, Mark},
  journal={Transactions of the Association for Computational Linguistics},
  volume={14},
  pages={1344--1368},
  year={2026},
  publisher={MIT Press 255 Main Street, 9th Floor, Cambridge, Massachusetts 02142, USA~…}
}

@article{bereska2024mechanistic,
  title={Mechanistic interpretability for AI safety--a review},
  author={Bereska, Leonard and Gavves, Efstratios},
  journal={arXiv preprint arXiv:2404.14082},
  year={2024}
}

@article{sharkey2025open,
  title={Open problems in mechanistic interpretability},
  author={Sharkey, Lee and Chughtai, Bilal and Batson, Joshua and Lindsey, Jack and Wu, Jeff and Bushnaq, Lucius and Goldowsky-Dill, Nicholas and Heimersheim, Stefan and Ortega, Alejandro and Bloom, Joseph and others},
  journal={arXiv preprint arXiv:2501.16496},
  year={2025}
}

@inproceedings{chang2024localization,
  title={Do localization methods actually localize memorized data in LLMs? a tale of two benchmarks},
  author={Chang, Ting-Yun and Thomason, Jesse and Jia, Robin},
  booktitle={Proceedings of the 2024 Conference of the North American Chapter of the Association for Computational Linguistics: Human Language Technologies (Volume 1: Long Papers)},
  pages={3190--3211},
  year={2024}
}

@article{cohen2024evaluating,
  title={Evaluating the ripple effects of knowledge editing in language models},
  author={Cohen, Roi and Biran, Eden and Yoran, Ori and Globerson, Amir and Geva, Mor},
  journal={Transactions of the Association for Computational Linguistics},
  volume={12},
  pages={283--298},
  year={2024},
  publisher={MIT Press One Broadway, 12th Floor, Cambridge, Massachusetts 02142, USA~…}
}

@article{harle2026measuring,
  title={Measuring and guiding monosemanticity},
  author={H{\"a}rle, Ruben and Friedrich, Felix and Brack, Manuel and Deiseroth, Bj{\"o}rn and Waeldchen, Stephan and Schramowski, Patrick and Kersting, Kristian},
  journal={Advances in Neural Information Processing Systems},
  volume={38},
  pages={167468--167508},
  year={2026}
}

@inproceedings{ferrando2025know,
  title={Do i know this entity? knowledge awareness and hallucinations in language models},
  author={Ferrando, Javier and Obeso, Oscar and Rajamanoharan, Senthooran and Nanda, Neel},
  booktitle={International Conference on Learning Representations},
  volume={2025},
  pages={77965--78000},
  year={2025}
}

@article{shi2025safer,
  title={Safer: Probing safety in reward models with sparse autoencoder},
  author={Shi, Wei and Xie, Ziyuan and Li, Sihang and Wang, Xiang},
  journal={arXiv preprint arXiv:2507.00665},
  year={2025}
}

@article{diehl2011hematopoietic,
  title={Hematopoietic cell types: prototype for a revised cell ontology},
  author={Diehl, Alexander D and Augustine, Alison Deckhut and Blake, Judith A and Cowell, Lindsay G and Gold, Elizabeth S and Gondr{\'e}-Lewis, Timothy A and Masci, Anna Maria and Meehan, Terrence F and Morel, Penelope A and Nijnik, Anastasia and others},
  journal={Journal of biomedical informatics},
  volume={44},
  number={1},
  pages={75--79},
  year={2011},
  publisher={Elsevier}
}

@article{soroush2024large,
  title={Large language models are poor medical coders—benchmarking of medical code querying},
  author={Soroush, Ali and Glicksberg, Benjamin S and Zimlichman, Eyal and Barash, Yiftach and Freeman, Robert and Charney, Alexander W and Nadkarni, Girish N and Klang, Eyal},
  journal={Nejm Ai},
  volume={1},
  number={5},
  pages={AIdbp2300040},
  year={2024},
  publisher={Massachusetts Medical Society}
}

@article{sun2024large,
  title={Are large language models a good replacement of taxonomies?},
  author={Sun, Yushi and Xin, Hao and Sun, Kai and Xu, Yifan Ethan and Yang, Xiao and Dong, Xin Luna and Tang, Nan and Chen, Lei},
  journal={arXiv preprint arXiv:2406.11131},
  year={2024}
}

@article{hou2024assessing,
  title={Assessing GPT-4 for cell type annotation in single-cell RNA-seq analysis},
  author={Hou, Wenpin and Ji, Zhicheng},
  journal={Nature methods},
  volume={21},
  number={8},
  pages={1462--1465},
  year={2024},
  publisher={Nature Publishing Group US New York}
}

@article{giunchiglia2022deep,
  title={Deep learning with logical constraints},
  author={Giunchiglia, Eleonora and Stoian, Mihaela Catalina and Lukasiewicz, Thomas},
  journal={arXiv preprint arXiv:2205.00523},
  year={2022}
}

@inproceedings{li2019augmenting,
  title={Augmenting neural networks with first-order logic},
  author={Li, Tao and Srikumar, Vivek},
  booktitle={Proceedings of the 57th annual meeting of the association for computational linguistics},
  pages={292--302},
  year={2019}
}

@article{marra2024statistical,
  title={From statistical relational to neurosymbolic artificial intelligence: A survey},
  author={Marra, Giuseppe and Duman{\v{c}}i{\'c}, Sebastijan and Manhaeve, Robin and De Raedt, Luc},
  journal={Artificial intelligence},
  volume={328},
  pages={104062},
  year={2024},
  publisher={Elsevier}
}

@inproceedings{xu2018semantic,
  title={A semantic loss function for deep learning with symbolic knowledge},
  author={Xu, Jingyi and Zhang, Zilu and Friedman, Tal and Liang, Yitao and Broeck, Guy},
  booktitle={International conference on machine learning},
  pages={5502--5511},
  year={2018},
  organization={PMLR}
}

@article{bussmann2025learning,
  title={Learning multi-level features with matryoshka sparse autoencoders},
  author={Bussmann, Bart and Nabeshima, Noa and Karvonen, Adam and Nanda, Neel},
  journal={arXiv preprint arXiv:2503.17547},
  year={2025}
}

@inproceedings{mudide2025efficient,
  title={Efficient dictionary learning with switch sparse autoencoders},
  author={Mudide, Anish and Engels, Josh and Michaud, Eric and Tegmark, Max and Schroeder de Witt, Christian},
  booktitle={International Conference on Learning Representations},
  volume={2025},
  pages={101830--101844},
  year={2025}
}

@inproceedings{bollacker2008freebase,
  title={Freebase: a collaboratively created graph database for structuring human knowledge},
  author={Bollacker, Kurt and Evans, Colin and Paritosh, Praveen and Sturge, Tim and Taylor, Jamie},
  booktitle={Proceedings of the 2008 ACM SIGMOD international conference on Management of data},
  pages={1247--1250},
  year={2008}
}

@inproceedings{speer2017conceptnet,
  title={Conceptnet 5.5: An open multilingual graph of general knowledge},
  author={Speer, Robyn and Chin, Joshua and Havasi, Catherine},
  booktitle={Proceedings of the AAAI conference on artificial intelligence},
  volume={31},
  number={1},
  year={2017}
}
